\documentclass{article}

\usepackage{microtype}
\usepackage{graphicx}
\usepackage{subfigure}
\usepackage{booktabs} 

\usepackage{hyperref}
\usepackage{xcolor}

\usepackage[accepted]{icml2021}

\usepackage{amsmath}
\usepackage{wrapfig}
\usepackage{graphicx}
\usepackage{subfigure}
\usepackage{color}
\usepackage{CJKutf8}
\usepackage{appendix}
\usepackage{multirow}
\usepackage{caption}
\usepackage{mathrsfs}

\icmltitlerunning{Temporally Correlated Task Scheduling for Sequence Learning}

\usepackage{amsmath,amsfonts,bm}

\def\eqref#1{equation~\ref{#1}}

\def\1{\bm{1}}

\DeclareMathAlphabet{\mathsfit}{\encodingdefault}{\sfdefault}{m}{sl}
\SetMathAlphabet{\mathsfit}{bold}{\encodingdefault}{\sfdefault}{bx}{n}

\DeclareMathOperator*{\argmax}{arg\,max}
\DeclareMathOperator*{\argmin}{arg\,min}

\newcommand{\domX}{\mathcal{X}}
\newcommand{\domY}{\mathcal{Y}}
\newcommand{\Ddev}{D_{\text{v}}}
\newcommand{\Dtrain}{D_{\text{t}}}
\newcommand{\waitk}{\textit{wait}-$k$}
\newcommand{\waitm}{\textit{wait}-$m$}

\newcommand{\Waitk}{\textit{Wait}-$k$}
\newcommand{\waitkstar}{\textit{wait}-$k^*$}
\newcommand{\Waitkstar}{\textit{Wait}-$k^*$}
\newcommand{\attn}{\texttt{attn}}

\newcommand{\red}[1]{{\color{red} #1}}

\newcommand{\markr}[1]{{\color{green} #1}}
\newcommand{\correct}[0]{{\color{green} green}}

\newcommand{\markw}[1]{{\color{red} #1}}
\newcommand{\wrong}[0]{{\color{red} red}}

\newcommand{\marka}[1]{{\color{blue} #1}}
\newcommand{\markb}[1]{{\color{cyan} #1}}
\newcommand{\markc}[1]{{\color{yellow} #1}}
\newcommand{\markd}[1]{{\color{magenta} #1}}

\newcommand{\domT}{\mathcal{T}}
\begin{document}

\twocolumn[
\icmltitle{Temporally Correlated Task Scheduling for Sequence Learning}



\icmlsetsymbol{equal}{*}

\begin{icmlauthorlist}
\icmlauthor{Xueqing Wu}{ustc}
\icmlauthor{Lewen Wang}{msra}
\icmlauthor{Yingce Xia}{msra}
\icmlauthor{Weiqing Liu}{msra}
\icmlauthor{Lijun Wu}{msra}
\icmlauthor{Shufang Xie}{msra}
\icmlauthor{Tao Qin}{msra}
\icmlauthor{Tie-Yan Liu}{msra}
\end{icmlauthorlist}

\icmlaffiliation{ustc}{University of Science and Technology of China, Hefei, Anhui, China}
\icmlaffiliation{msra}{Microsoft Research, Beijing, China}

\icmlcorrespondingauthor{Yingce Xia}{yingce.xia@microsoft.com}

\icmlkeywords{Simultaneous machine translation, stock trend forecasting}

\vskip 0.3in
]



\printAffiliationsAndNotice{}  

\begin{abstract}
Sequence learning has attracted much research attention from the machine learning community in recent years. In many applications, a sequence learning task is usually associated with multiple temporally correlated auxiliary tasks, which are different in terms of how much input information to use or which future step to predict. For example, (i) in simultaneous machine translation, one can conduct translation under different latency (i.e., how many input words to read/wait before translation); (ii) in stock trend forecasting, one can predict the price of a stock in different future days (e.g., tomorrow, the day after tomorrow). While it is clear that those temporally correlated tasks can help each other, there is a very limited exploration on how to better leverage multiple auxiliary tasks to boost the performance of the main task. In this work, we introduce a learnable scheduler to sequence learning, which can adaptively select auxiliary tasks for training depending on the model status and the current training data. The scheduler and the model for the main task are jointly trained through bi-level optimization. Experiments show that our method significantly improves the performance of simultaneous machine translation and stock trend forecasting. 
\end{abstract}

\section{Introduction}\label{sec:introduction}
Sequence learning~\cite{NIPS2014_5346,bahdanau2014neural} is an important problem in deep learning, which covers many applications including machine translation~\citep{wu2016google,vaswani2017attention}, time series prediction~\citep{ALSTM,SFM}, weather forecasting~\cite{10.5555/2969239.2969329,DBLP:journals/corr/abs-1711-02316}, etc. In real-world applications, a sequence learning task is often associated with multiple temporally correlated tasks: (1) In simultaneous machine translation, we need to begin translation before reading the complete source sequence. In this application, the generation of output sentence is $k$ word behind the source input, which is called \waitk{} strategy/task~\citep{ma2019stacl}. \Waitk{} tasks with different $k$ values are temporally correlated. (2) Stock forecasting aims to predict the trend of stocks at some future day~\cite{dai2013machine}. Tasks of predicting stock prices at different future days constitute a set of temporally correlated tasks. Different from general multi-task learning, the input and output data of these temporally correlated tasks are highly overlapped, and only differ in how much future input to use when processing the current input (e.g., how many future source words to read to translate the current word in simultaneous translation) or which future data to predict (e.g., the price movement of which future day to predict in stock trend forecasting). 

Clearly, a given sequence learning task (denoted as the main or target task) can be boosted by its temporally correlated tasks. Taking English-to-Vietnamese simultaneous translation as an example, assume the main task is \textit{wait}-$3$, and the temporally correlated tasks are \textit{wait}-$1,2,3,\cdots,13$. By training a model on all those correlated tasks 
and evaluating the model on \textit{wait}-$3$, we observe that the BLEU score is improved from $27.17$ to $28.54$. (BLEU~\cite{papineni-etal-2002-bleu} is a widely used accuracy metric in machine translation, the larger the better.) A similar phenomenon is observed in stock price forecasting. Improvements across various applications motivate us to explore how to leverage those temporally correlated tasks more effectively to improve performance on the main task.

Given that there are usually multiple temporally correlated tasks, the key challenge lies in which tasks to use and when to use them in the training process. In this work, we introduce a learnable task scheduler for sequence learning, which adaptively selects temporally correlated tasks during the training process. The scheduler accesses the model status and the current training data (e.g., in current minibatch),
and selects the best auxiliary task to help the training of the main task. The scheduler and the model for the main task are jointly trained through bi-level optimization: the scheduler is trained to maximize the validation performance of the model, and the model is trained to minimize the training loss guided by the scheduler. Experimental results on simultaneous machine translation and stock trend forecasting show that such a task scheduler brings significant improvements.

Our main contributions are three-fold:

\noindent(1) To the best of our knowledge, we are the first to formulate and study the problem of leveraging temporally correlated tasks for sequence learning.

\noindent(2) We propose a new learning algorithm with a scheduler to adaptively select the temporally correlated tasks to boost the main task. The scheduler and the model of the main task are jointly trained without additional supervision signals.

\noindent(3) We empirically verify the effectiveness of our method. Experiments on four simultaneous translation tasks show that our method improves \waitk{} baseline by $1$ to $3$ BLEU scores. Experiments on a stock price forecasting dataset show that our method can improve the baselines in terms of Spearman's rank correlation coefficient and prediction loss.

Our code for simultaneous translation task is  at \url{https://github.com/shirley-wu/simul-mt_temporally-correlated-task-scheduling}, and the code for stock price forecasting is  at\newline \url{https://github.com/microsoft/qlib/tree/main/examples/benchmarks/TCTS}.

\section{Problem Setup and Our Algorithm}~\label{sec:framework}
In this section, we first set up the problem of temporally correlated task scheduling for sequence learning, and then present our algorithm. Finally, we discuss the connection between our method and existing methods, and also provide an intuition why our method works.

\subsection{Problem Setup}\label{sec:problem:Setup}
Let $\domX$ and $\domY$ denote the input space and output space, and 
let $\domT=\{T_1,\cdots,T_{|\domT|}\}$ denote a set of temporally correlated tasks. Depending on applications, those tasks can be temporally correlated in different ways. We categorize them into two main classes and formalize them as follows. 

(1) {\em Sequence processing with latency constraints}: Tasks in this category aim to process each token in an input sequence, one by one sequentially, under certain latency constraints. Representative examples include simultaneous translation~\cite{ma2019stacl}, which targets at translating each source word without waiting for the end of the input sentence, and streaming speech recognition~\cite{8682336}, which requires recognizing each word said by a speaker in real-time, etc.

Let $x$ denote an input sequence and $x_t$ denotes its $t$-th token. We focus on the setting where each token in $x$ will be sequentially processed. Each task $T\in\domT$ is associated with a non-negative integer $\delta_T$, indicating that when processing the token $x_t$, how many additional tokens can be used. Note that in conventional sequence learning where the complete source sequence is available, $\delta_T=\infty$. In the extreme case where no latency is allowed, $\delta_T=0$. Let $f_T$ denote the model for task $T$, which generates the output sequentially. $\tilde{y}$ denotes the output sequence of $f_T$ based on $x$, and is initialized as an empty list. After processing $x_t$ conditioned on $x_{\le t + \delta_T}=(x_1,x_2,\cdots,x_{t+\delta_T})$, based on the processed representations of $x_t$, $x_{\le t + \delta_T}$, and the already generated $\tilde{y}$, $f_T$ needs to make a decision $\hat{y}=f_T(x_t, x_{\le t + \delta_T}, \tilde{y})$. The decision could be to generate a single token, multiple tokens, or no token. After that, we append $\hat{y}$ to the end of $\tilde{y}$. We repeat the above steps to process all tokens in $x$.

(2) \emph{Predicting the future}: Tasks in this category aim to predict some future values for a given time series. Representative examples include stock trend forecasting~\cite{dai2013machine}, where we want to predict the price trend of an individual stock on some future days (e.g., tomorrow, the day after tomorrow, etc.) based on its historical prices, and weather forecasting~\cite{10.5555/2969239.2969329,DBLP:journals/corr/abs-1711-02316}, where we want to predict the weather of a city on a future day based on its historical weather.

Let $x$ denote the input time series, and $y=(y_1,y_2,\cdots)$ denote the output sequence, where $y_t$ is the value of the $t$-th future day/step. For a task $T\in\domT$, the goal is to predict the value of the $\sigma_T$-th future day, i.e., $\tilde{y}_{\sigma_T}$, where $\sigma_T$ is a positive integer and varies across different tasks $T$.

Each task has a different loss function $\ell(x,y,T)$, which is used to measure the distance between the predicted $\tilde{y}$ and the groundtruth $y$ (in the first class of tasks) or $\tilde{y}_{\sigma_T}$ and $y_{\sigma_T}$ (in the second class of tasks).
Any task in $\domT$ can be the main task, and the others are regarded as auxiliary tasks. In this work, our goal is to better leverage those auxiliary tasks to help the main task.

\begin{figure}[!t]
\centering
\includegraphics[width=0.8\linewidth]{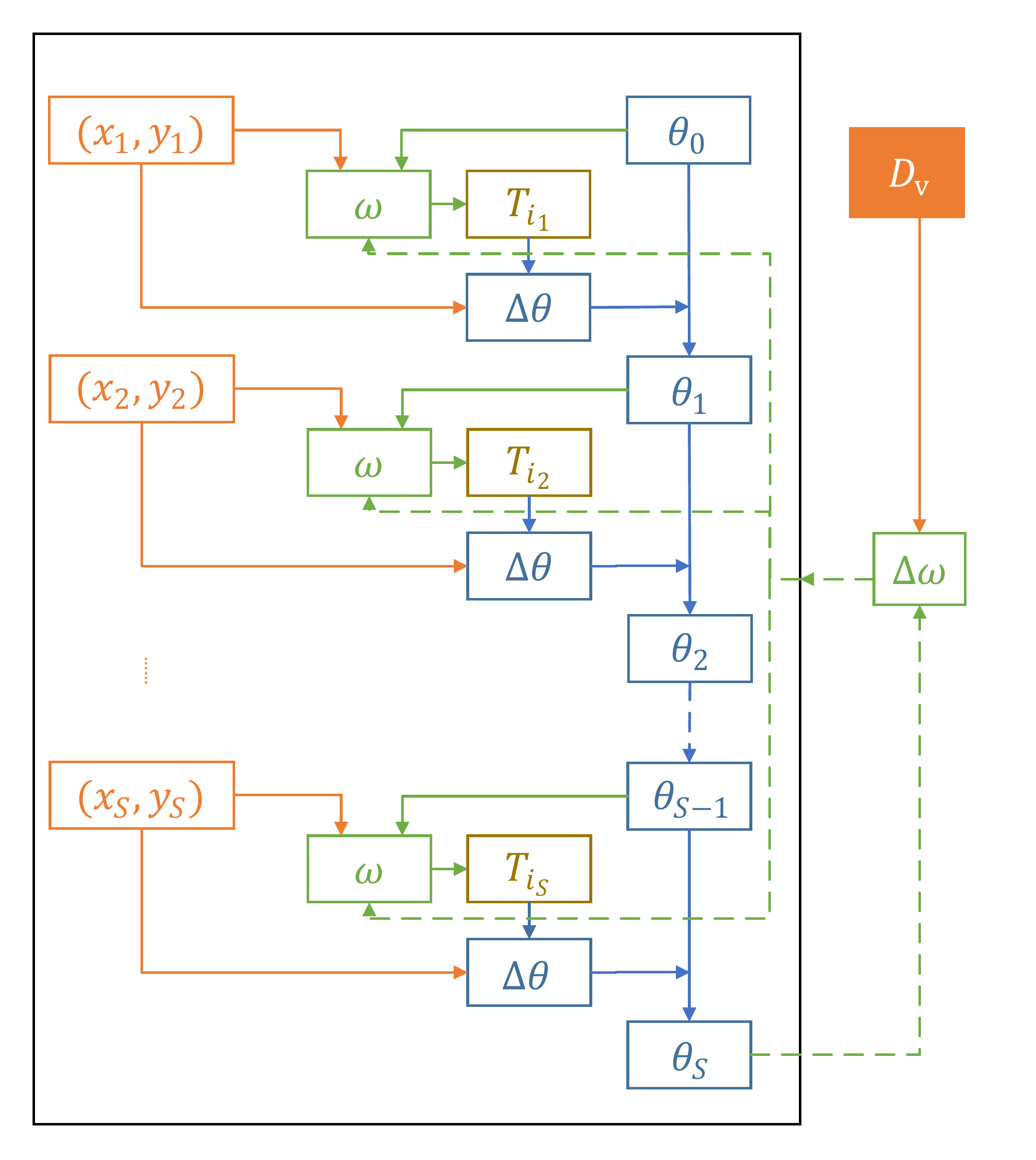}
\caption{The optimization workflow of one episode. The solid lines and dashed lines represent the workflow of inner-optimization ({\em w.r.t.} $\theta$) and outer-optimization ({\em w.r.t.} $\omega$) respectively. At  step $s$, with training data $(x_s,y_s)$, the scheduler $\varphi(\cdots;\omega)$ chooses a suitable task $T_{i_s}$ (green solid lines) to update the model $f(\cdots;\theta)$ (blue solid lines). After $S$ steps, we evaluate the model $f$ on the validation set $\Ddev$ and update the scheduler $\varphi$ (green dashed lines).}
\label{fig:algo_illustration}
\end{figure}

\subsection{Our Algorithm}
As aforementioned, our basic idea is to introduce a task scheduler, which adaptively selects an auxiliary training task to improve the performance of the main task. A basic assumption of our algorithm is that the models for all tasks in $\domT$ can share parameters (except for a few individual parts like task embeddings), which is easy to satisfy in machine translation, stock trend forecasting, etc. Let $f(x;\theta)$ denote the model with parameter $\theta$. Let $\varphi(x,y,\theta;\omega)$ denote the task scheduler parameterized by $\omega$, which outputs an auxiliary task in $\domT$ conditioned on an input-output pair $(x,y)$ and the model $f(\cdots;\theta)$ of the main task.
More specifically, we extract a low-dimensional feature $I_{x,y,\theta}$ to describe $(x,y,\theta)$ and use it as the actual input of the scheduler network $\varphi$. For ease of reference, we will alternatively use $\varphi(x,y,\theta;\omega)$ and $\varphi(I_{x,y,\theta};\omega)$.
$\varphi$ can be different functions depending on applications, which will be discussed in Section~\ref{sec:simMT} and \ref{sec:stock}.

Now we have two sets of parameters to learn from the data: $\omega$ in the scheduler $\varphi$  and $\theta$ in the model $f$. Let $\Dtrain$ and $\Ddev$ denote the training and validation sets respectively. 
We formulate the training of $f$ and $\varphi$ as a bi-level optimization~\cite{sinha2017review}, which is
\begin{align}
& \max_{\omega} \mathcal{M}(\Ddev;\theta(\omega)); \quad {\rm s.t.},\label{eq:high_level_obj_scheduler}\\
&\theta(\omega)=\argmin_{\theta} \frac{1}{\vert\Dtrain\vert}\sum_{(x,y)\sim\Dtrain}\mathbb{E}_{T\sim\varphi(x,y,\theta;\omega)}\ell(x,y,T;\theta). \nonumber
\end{align}
In the above formulation, $\mathcal{M}(\Ddev;\theta(\omega))$ denotes the validation performance of the main task, and $\ell$ is the loss function of data $(x,y)$ on task $T$. Intuitively, we expect that the scheduler can select a good auxiliary task for each input-output pair, so that eventually the validation performance of the model on the main task can be maximized. 

Following the common practice~\cite{liu2018darts,l2t2019yang}, we optimize Eqn.(\ref{eq:high_level_obj_scheduler}) in an alternative way: We first fix $\omega$ and optimize $\theta$, and then fix $\theta$ and optimize $\omega$ using the REINFORCE algorithm~\cite{Sutton1998}. We repeat the process until convergence, as shown in Algorithm~\ref{alg:optimizer_scheduler}.
The training process consists of $E$ episodes (i.e., the outer loop), and each episode consists of $S$ update iterations (i.e., the inner loop). The inner loop (from step 4 to step 9) aims to optimize $\theta$, where we can update the parameter with any gradient-based optimizer like momentum SGD, Adam~\citep{adam_optimizer}, etc.
The outer loop (from step 2 to step 11) aims to optimize $\omega$. $\varphi(x,y,\theta;\omega)$ can be regarded as a policy network, where the state is the feature $I_{x,y,\theta}$,  the action is the choice of task $T\in\domT$, and the reward is the validation performance $R_e$ (step 10). At the end of each episode, we update $\omega$ using REINFORCE algorithm (step 11).
The workflow of one episode is shown in Figure~\ref{fig:algo_illustration}.

\begin{algorithm}[tb]
\small
\caption{The optimization algorithm.}
\label{alg:optimizer_scheduler}
\begin{algorithmic}[1]
\STATE {\bfseries Input:} Training episode $E$; internal update iterations $S$; learning rate $\eta_1$ of model $f(\cdots;\theta)$ ; learning rate $\eta_2$ of the scheduler $\varphi(\cdots;\omega)$; batch size $B$; 
\FOR{$e\leftarrow 1:E$}
\STATE Prepare a buffer to store states and actions: $\mathcal{B}=\{\}$; 
\FOR{$s\leftarrow 1:S$}
\STATE Sample a minibatch of data $D_{e,s}$ from $\Dtrain$ with size $B$;
\STATE For each $(x,y)\in D_{e,s}$, we sample a temporally correlated auxiliary task $T$ according to the output distribution of $\varphi(x,y,\theta;\omega)$, and get $\tilde{D}=\{(x,y,T)\}$ where $|\tilde{D}|=B$;
\STATE  Add the inputs of $\varphi$ (i.e., $I_{x,y,\theta}$) to $\mathcal{B}$:\\
$\mathcal{B}\leftarrow\mathcal{B}\cup \{(I_{x,y,\theta},T)|(x,y,T) \in \tilde{D}\}$;
\STATE Update model $f$: \\
$\theta\leftarrow\theta - (\eta_1 / B) \nabla_\theta\sum_{(x,y,T)\in\tilde{D}}\ell(x,y,T;\theta)$;
\ENDFOR
\STATE Calculate the validation performance as the reward: $R_e = \mathcal{M}(\Ddev;\theta_{e,S})$;
\STATE Update the scheduler: \\
$\omega\leftarrow\omega + \eta_2 R_e \sum_{(I,T)\in\mathcal{B}}\nabla_\omega\log P(\varphi(I;\omega)=T)$.
\ENDFOR
\STATE {\bfseries Return:} $\theta$.
\end{algorithmic}
\end{algorithm}

\subsection{Discussion}\label{sec:alg_part:discussion}

Our problem is related to several other learning settings.

\noindent(1) In multi-task learning~\cite{zhang2017multi_task_survey} (briefly, MTL), multiple tasks are jointly trained and mutually boosted. In MTL, each task is treated equally, while in our setting, we focus on the main task. 

\noindent(2) Transfer learning~\cite{Pan10asurvey} also leverages auxiliary tasks to boost a main task. Curriculum transfer learning (briefly, CL)~\cite{shao2018starcraft,dong2017multi} is one kind of transfer learning which schedules auxiliary tasks according to certain rules, e.g., from easy  to difficult. Our problem can also be regarded as a special kind of transfer learning, where the auxiliary tasks are temporally correlated with the main task.  Our learning process is dynamically controlled by a scheduler rather than some pre-defined rules.

\noindent(3) Our method is also related to Different Data Selection (DDS) \citep{wang2020optimizing}. Both our work and DDS train a meta controller to improve the learning process on a given task via reinforcement learning. The main difference is that DDS aims to better select training data from a dataset, while our work aims to better utilize temporally correlated tasks (and their corresponding data). Given a main task, our method constructs and schedules a series of temporal correlated tasks, while DDS works on the given task only.

Intuitively, training with temporally correlated auxiliary tasks  helps the main task for two reasons: (1) The temporally correlated tasks either share the same output with overlapped inputs or the same input. Thus, joint training of them enables shared representation learning, just like general multi-task training (MTL), and consequently improves the main task; (2) The $x/y$ pairs from correlated tasks can be viewed as noised augmentation of the training data for the main task, which helps the main task.

\section{Application to Simultaneous Translation}\label{sec:simMT}

\subsection{Task Description}
Simultaneous neural machine translation (briefly, NMT) has attracted much attention recently. In contrast to standard NMT, where the NMT system can access the full input sentence, simultaneous NMT requires the model to conduct translation before reading the full input sentence. In this task, $\domX$ and $\domY$ are collections of sentences in the source language and the target language. 
\Waitk{}~\citep{ma2019stacl} is a simple yet effective method in simultaneous NMT, where the translation is $k$ words behind the source input. Formally, given a data pair $(x,y)\in\domX\times\domY$ where the $t$-th token of $x$/$y$ is $x_t$/$y_t$, when we read a prefix $(x_1,x_2,\cdots,x_{k})$ (briefly denoted as $x_{\le k}$), we need to output the first target token $y_1$; when reading $x_{\le k+1}$, we should output $y_2$, etc. In comparison, for standard NMT, one can start translation after reading all tokens in $x$. A larger $k$ makes the task easier and leads to better translation quality, but at the cost of a larger waiting time.

\iftrue
Under our framework, simultaneous translation belongs to the first class of tasks in Section ~\ref{sec:problem:Setup}. Let $T_m$ denote \waitm{} task, and $\delta_{T_m}$ for task $T_m$ is $m-1$.
Assuming \waitk~is the main task $T_{\rm main}$, then the temporally correlated tasks $\domT$ are \waitm{} with different $m$ values, i.e. $\domT=\{T_1,T_2,\cdots,T_M\}$.
\else
Assuming \waitk{} is the main task $T_{\rm main}$, then the temporally correlated tasks $\domT$ are \waitm{} with different $m$ values. Let $T_m$ denote the \waitm{} task (i.e., the $\delta_T$ in Section~\ref{sec:problem:Setup} is $m−1$), so that $\domT=\{T_1,T_2,\cdots,T_M\}$. 
\fi
In this work, we set $M=13$. If the scheduler $\varphi$ assigns a task $T_m$ (i.e., \waitm{}) to a data pair $(x,y)$, the loss function on this task is:
\begin{equation}
\ell(x,y,T_m;\theta) = \sum_{t=1}^{|y|} \log P(y_t|y_{\le t-1},x_{\le t+m-1}; \theta).
\end{equation}
\subsection{Background}\label{sec:simMT:background}

Previous work on simultaneous NMT can be categorized by whether using a fixed decoding strategy or an adaptive one.
Fixed strategies use pre-defined rules to determine when to read or to write a new token \citep{dalvi2018incremental,ma2019stacl}.
\Waitk~is a representative method ~\citep{ma2019stacl}, which achieves good results in terms of translation quality and controllable latency, and has been used in speech-related simultaneous NMT~\citep{zhang2019simuls2s,ren2020simulSpeech}. 

For  adaptive strategies, \citet{DBLP:journals/corr/ChoE16} proposed wait-if-worse (WIW) and wait-if-diff (WID) methods which generate a new target word if its probability does not decrease (for WIW) or if the generated word is unchanged (for WID) after reading a new source token. \citet{grissom-ii-etal-2014-dont} and \citet{gu-etal-2017-learning} used reinforcement learning to train the read/write controller, while \citet{zheng-etal-2019-simpler} trained the controller in a supervised way. \citet{alinejad-etal-2018-prediction} added a ``predict'' operator to the controller so that it can anticipate future source inputs. 
\citet{zheng-etal-2019-simultaneous} introduced a ``delay'' token into the target vocabulary 
indicating that the model should read a new word. 
Monotonic infinite lookback attention (MILk) used a hard attention model to determine when to read new tokens, and a soft attention model to perform translation \cite{arivazhagan-etal-2019-monotonic}. 
\citet{ma2020monotonic} introduced multi-head attention to MILk and proposed monotonic multihead attention (MMA) with two variants: MMA-IL (Infinite Lookback) with higher translation quality and greater computational overhead, and MMA-H(ard) with higher computational efficiency.
Besides, \citet{zheng2020simultaneous} extended \waitk{} to an adaptive strategy by training multiple \waitm{} models with different $m$'s and adaptively selecting a decoding strategy during inference. 
\citet{zheng-etal-2020-opportunistic} explored a new setting, where at each timestep, the translation model over-generates the target words and corrects them in a timely fashion.

\subsection{Settings}\label{sec:experiments_simmt:settings}
\noindent{\em Datasets}:
For IWSLT'14 En$\to$De, following~\citep{edunov-etal-2018-classical}, we split $7k$ sentences from the training corpus for validation, and the test set is the concatenation of {\em tst2010, tst2011, tst2012, dev2010} and {\em dev2012}. For IWSLT'15 En$\to$Vi, following~\citep{ma2020monotonic}, we use {\it tst2012} as the validation set and {\it tst2013} as the test set. For IWSLT'17 En$\to$Zh, we concatenate {\it tst2013}, {\it tst2014} and {\it tst2015} as the validation set and use {\it tst2017} as the test set. For WMT'15 En$\to$De, following \citep{ma2019stacl, arivazhagan-etal-2019-monotonic}, we use {\it newstest2013} as the validation set and use {\it newstest2015} as the test set. More details about datasets can be found in Appendix C.1. 

\noindent{\em Models}: Following \citet{ma2019stacl}, the translation model $f$ is based on Transformer \cite{vaswani2017attention}. However, different from \citet{ma2019stacl}, we conduct incremental encoding, so that when encoding each source token, the model can only attend to its previous tokens.
The computation complexity for encoding is $O(L_x^2)$ compared to $O(L_x^3)$ in \citet{ma2019stacl}, where $L_x$ is the length of the source sentence $x$.
We find that our model is more efficient than \citet{ma2019stacl} with little accuracy drop.
Detailed formulation and empirical comparison are in Appendix B. 

For IWSLT En$\to$Zh and En$\to$Vi, we use the transformer small model, where the embedding dimension, feed-forward layer dimension, number of layers are $512$, $1024$ and $6$ respectively. For IWSLT En$\to$De, we use the same architecture but change the embedding dimension into $256$. For WMT'15 En$\to$De, we use the transformer big setting, where the above three numbers are $1024$, $4096$ and $6$ respectively.
The scheduler $\varphi$ for each task is a multilayer perceptron (MLP) with one hidden layer and the \texttt{tanh} activation function. The size of the hidden layer is $256$.

\noindent{\em Input features of $\varphi$}: The input of $\varphi$ is a $7$-dimension vector with the following features: (1) the ratios between the lengths of the source/target sentences to the average source/target sentence lengths in all training data ($2$ dimensions), i.e., $L_x/(\sum_{x'\in\domX}L_{x'}/|\domX|)$ and $L_y/(\sum_{y'\in\domY}L_{y'}/|\domY|)$; (2) the training loss over data $(x,y)$ evaluated on the main task \waitk; (3) the average of historical training losses; (4) the validation loss of the previous epoch; (5) the average of historical validation loss; (6) the ratio of current training step to the total training iteration. The ablation study of feature selection is in Appendix E.1. 

\noindent{\em Choice of $\mathcal{M}$}: The validation performance $\mathcal{M}$ is the negative validation loss with \waitk{} strategy. To stabilize training, the reward for the $e$-th episode (step 10 of Algorithm~\ref{alg:optimizer_scheduler}) is $R_e-R_{e-1}$, where $R_{e-1}$ is the validation performance $\mathcal{M}$ of the previous episode. 

\iftrue

\noindent{\em Baselines}: We implement the MTL and CL baselines discussed in Section~\ref{sec:alg_part:discussion}: (1) For MTL, each task is randomly sampled. (2) For the curriculum transfer learning (briefly, CL), we start from the easiest task $T_M$ and gradually move to the main task $T_{\rm main}$. For \waitk{}, we implement another variant \waitkstar{}, where we train $M$ baseline models on different \waitm{} tasks ($1\le m \le M$), and pick the best model according to the validation performance on \waitk{} task. \Waitkstar{} is expected to bring additional improvements as pointed out by \citet{ma2019stacl}. After that, we compare with several adaptive methods, including Wait-if-Worse (WIW), Wait-if-Diff (WID),  MILk, MMA-IL and MMA-H (refer to Section~\ref{sec:simMT:background} for a brief introduction). Finally, we combine our method with \citet{zheng2020simultaneous}.
We leave the training details of all algorithms (optimizer, hyperparameter, etc) in Appendix C.2, and the  details of baselines  in Appendix C.3.

\else

\noindent{\em Baselines}: We implement the MTL and CL baselines discussed in Section~\ref{sec:alg_part:discussion}. For \waitk{}, we implement another variant \waitkstar{}, where we train $M$ baseline models on $M$ different \waitm{} tasks ($1\le m \le M$), and pick the best model according to the validation performance on \waitk{} task. \Waitkstar{} is expected to bring additional improvement as pointed out by \citet{ma2019stacl}. After that, we compare with several adaptive methods, including Wait-if-Worse (WIW), Wait-if-Diff (WID),  MILk, MMA-IL and MMA-H (refer to Section~\ref{sec:simMT:background} for a brief introduction). Finally, we combine our method with \citet{zheng2020simultaneous}.
We leave the training details of all algorithms (optimizer, hyperparameter, etc) in Appendix C.2, and the  details of baselines  in Appendix C.3.

\fi

\noindent{\em Evaluation}: We use BLEU to measure the translation quality, and use Average Proportion (AP) and Average Lagging (AL) to evaluate the translation latency. AP measures the average proportion of source symbols required for translation, and AL measures the average number of delayed words 
(see Appendix A.1. for details).
Following the common practice~\citep{ma2019stacl, ma2020monotonic}, we show the BLEU-AP and BLEU-AL curves to demonstrate the tradeoff between quality and latency. For IWSLT'14 En$\to$De and IWSLT'15 En$\to$Vi, we use \texttt{multi-bleu.perl} to evaluate BLEU scores; for IWSLT'17 En$\to$Zh and WMT'15 En$\to$De, we use \texttt{sacreBLEU} to evaluate the detokenized BLEU scores.

\subsection{Results and Analysis}
\begin{figure}[!htb]
\centering
\subfigure[BLEU-AP, En$\to$Vi.]{\includegraphics[width=0.49\linewidth]{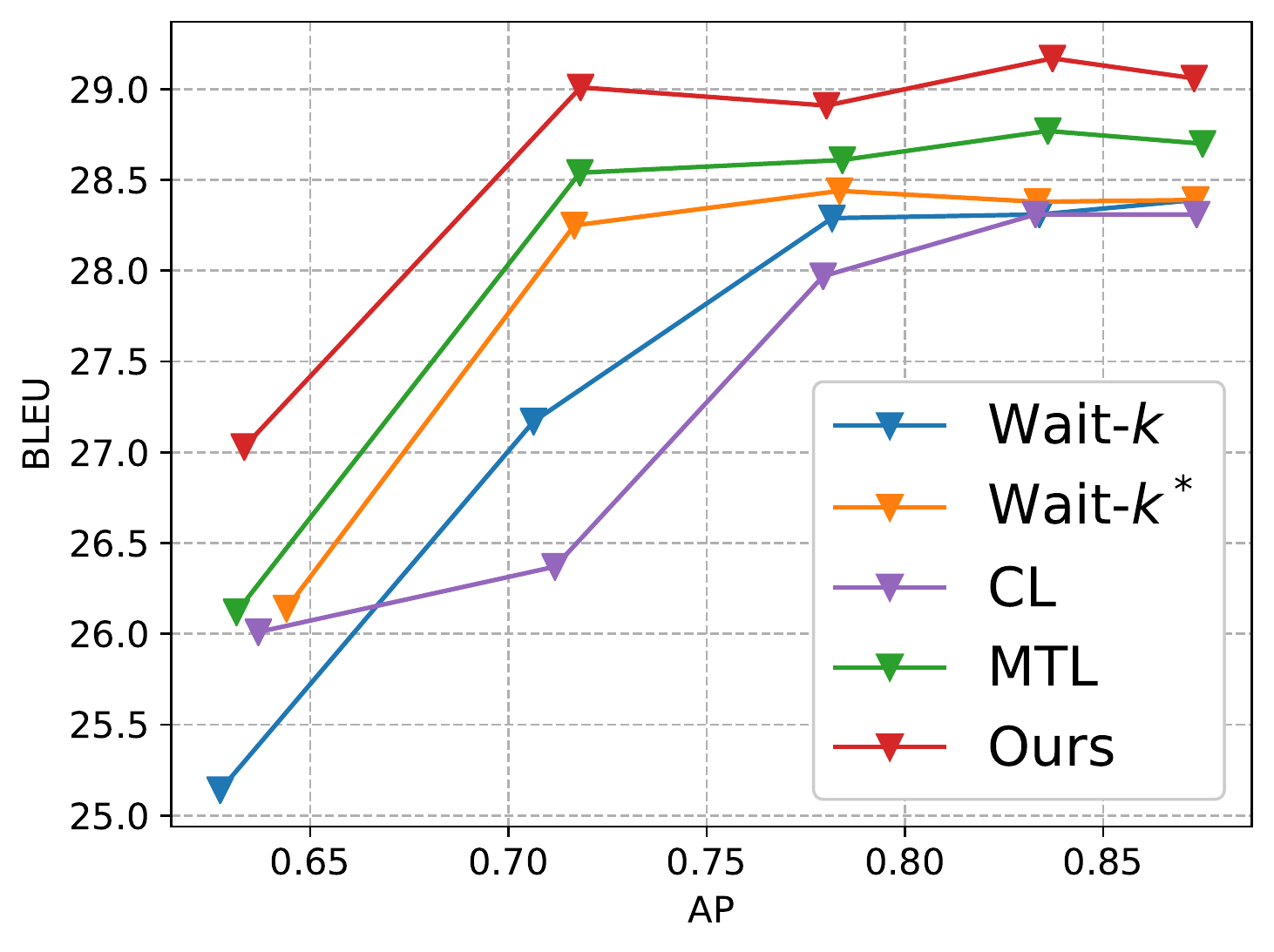}}
\subfigure[BLEU-AL, En$\to$Vi.]{\includegraphics[width=0.49\linewidth]{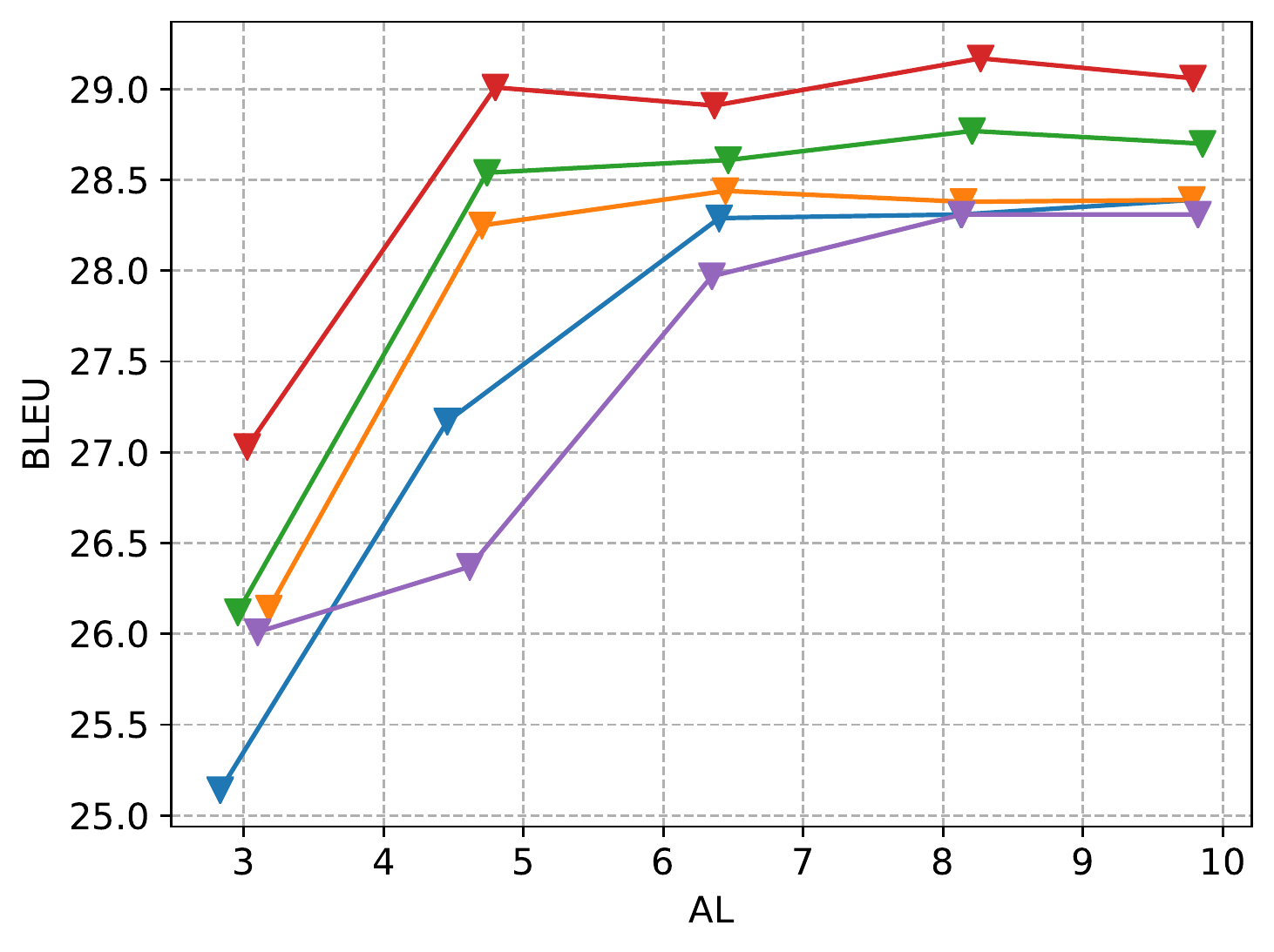}}

\subfigure[BLEU-AP, En$\to$De.]{\includegraphics[width=0.49\linewidth]{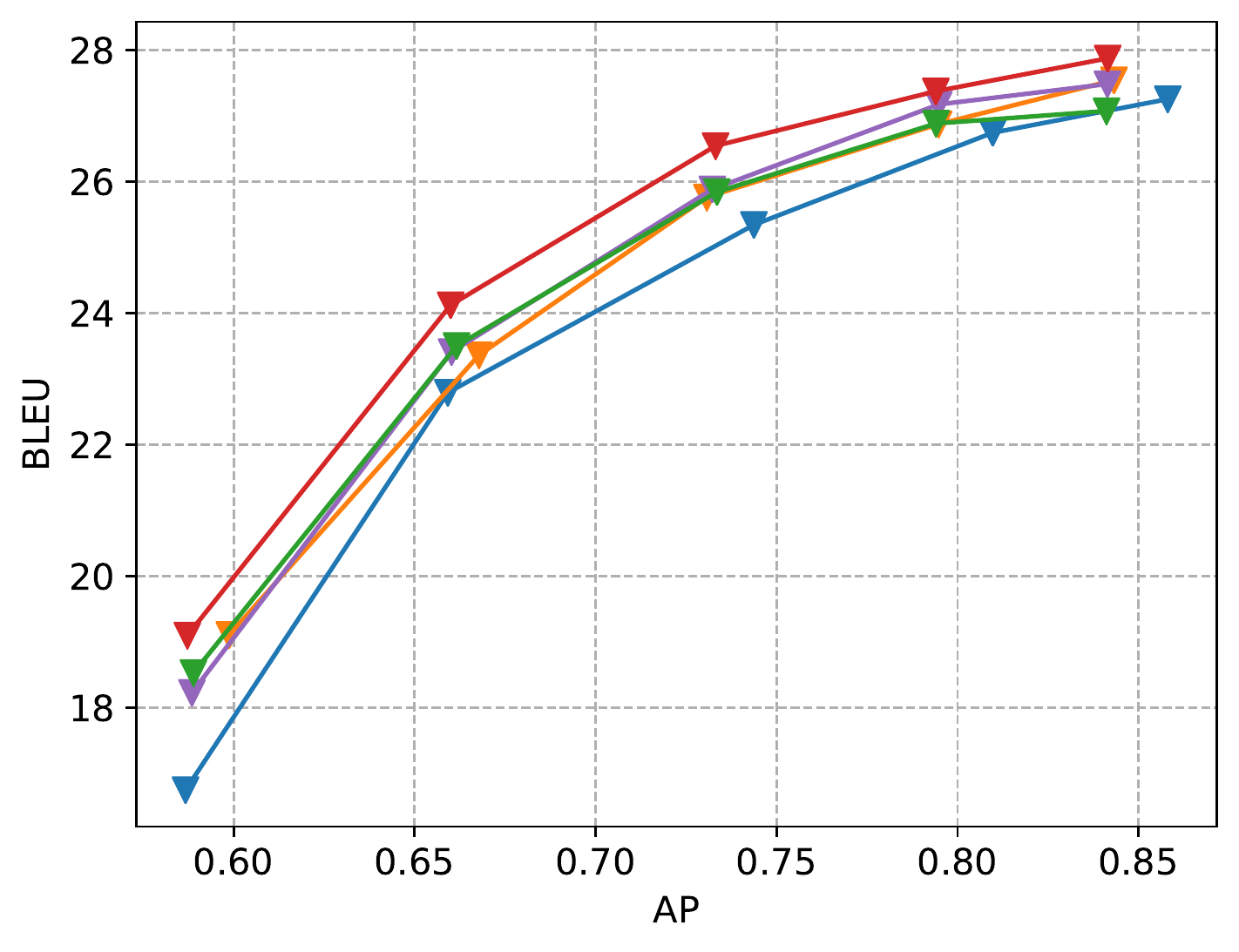}}
\subfigure[BLEU-AL, En$\to$De.]{\includegraphics[width=0.49\linewidth]{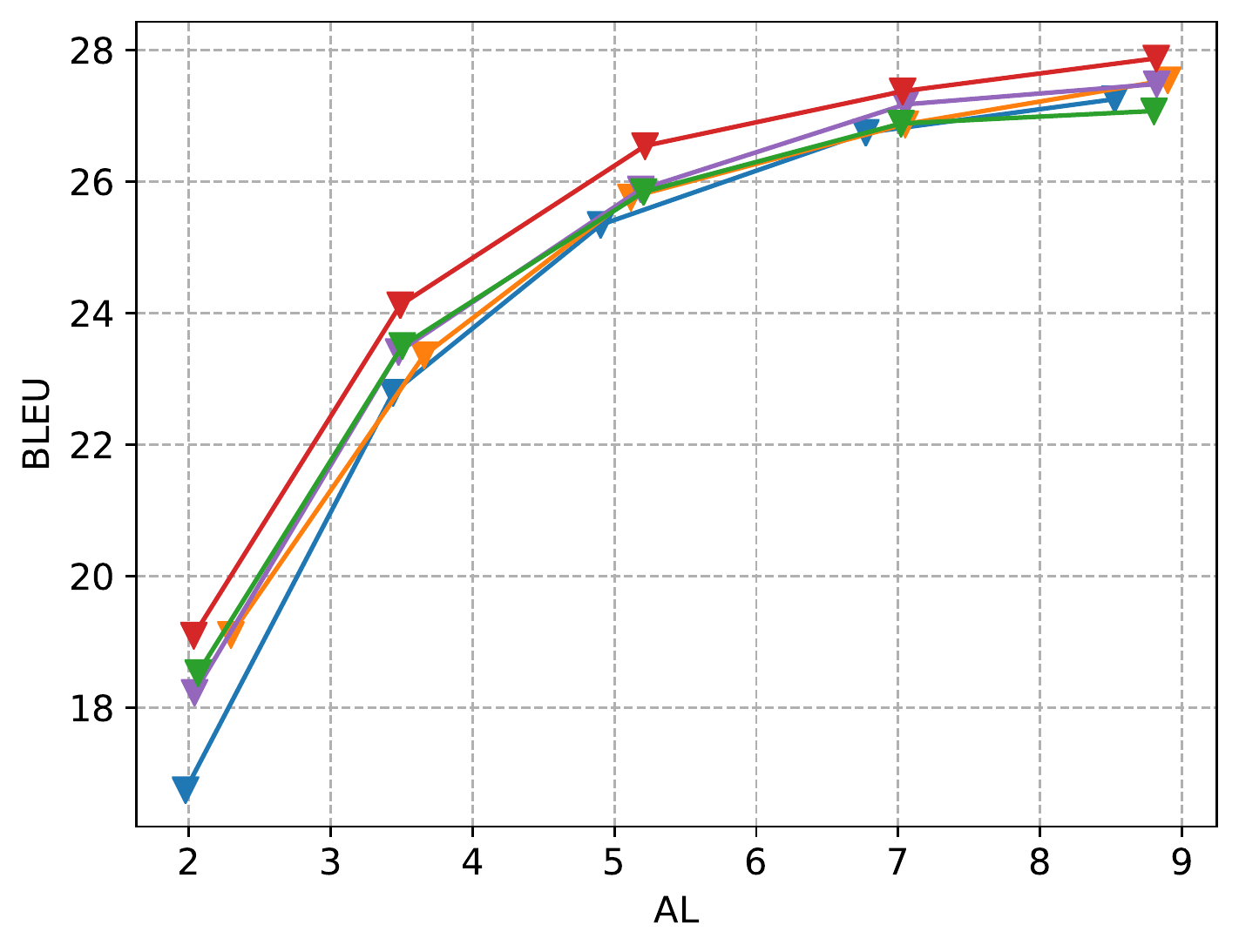}}

\subfigure[BLEU-AP, En$\to$Zh.]{\includegraphics[width=0.49\linewidth]{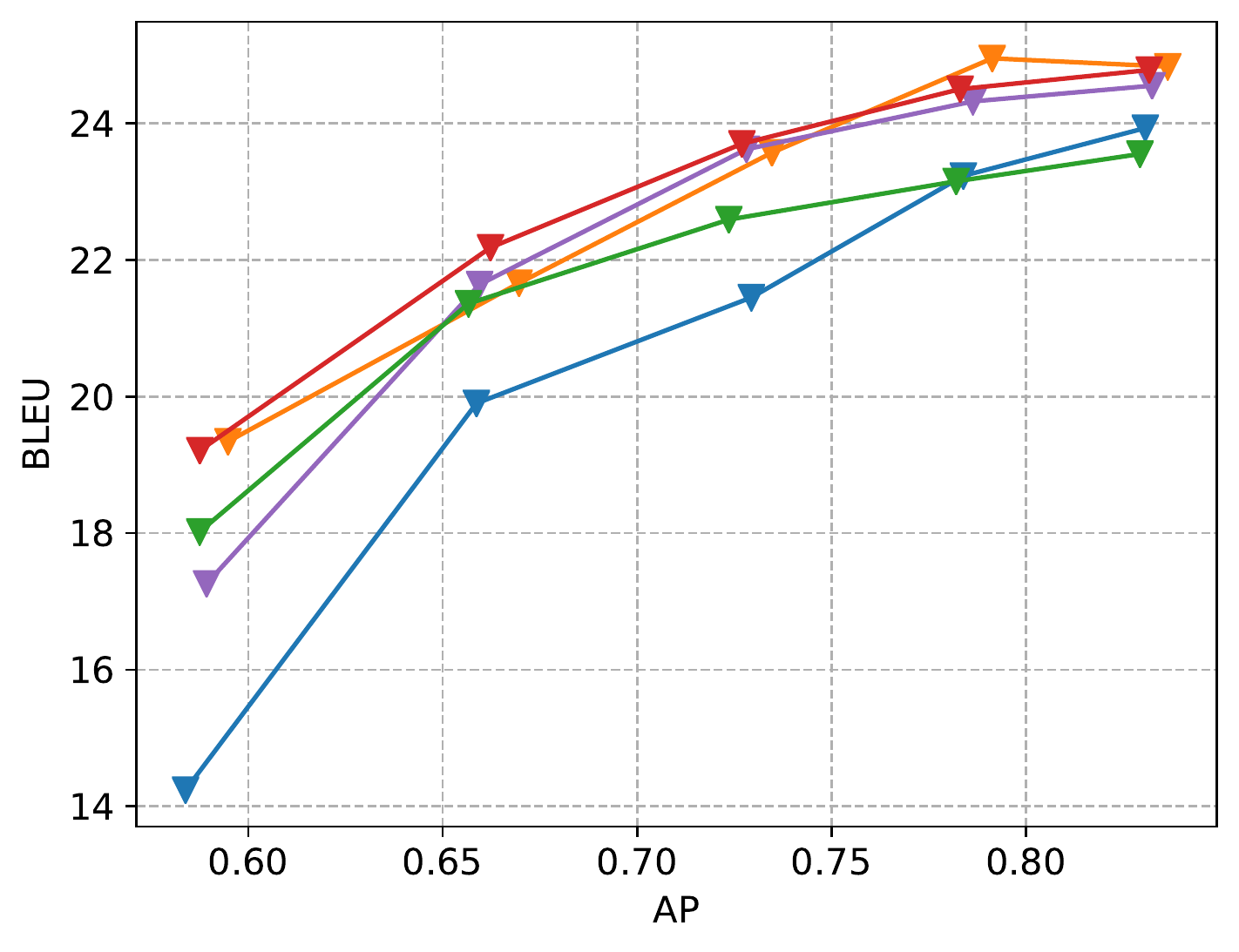}}
\subfigure[BLEU-AL, En$\to$Zh.]{\includegraphics[width=0.49\linewidth]{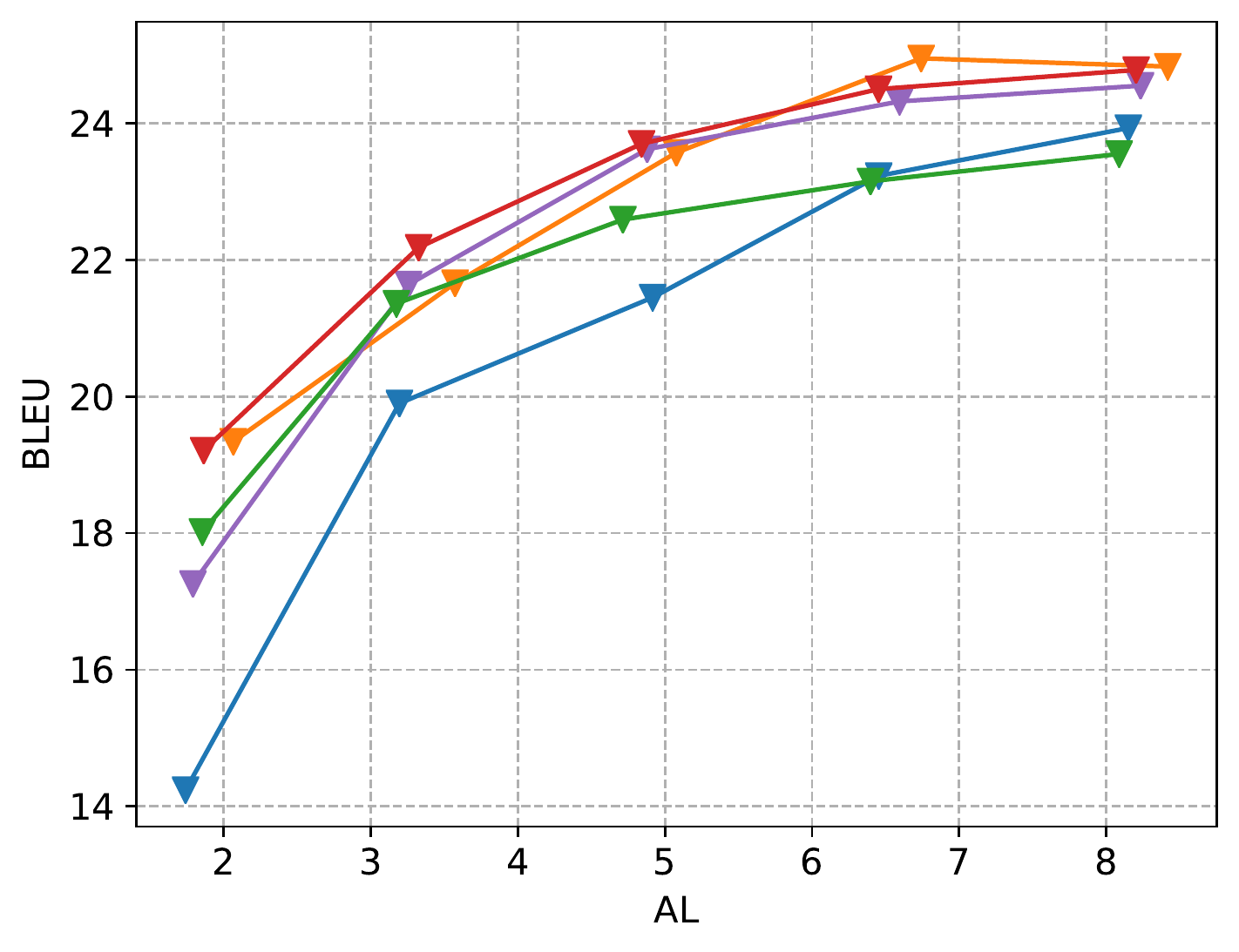}}
\caption{Translation quality against latency metrics (AP and AL) on IWSLT tasks.
}
\label{fig:iwslt_bleu_latency}
\end{figure}

We first compare our method with MTL, CL and \waitkstar{}  on IWSLT datasets. The BLEU-latency curves are shown in Figure~\ref{fig:iwslt_bleu_latency}, and the BLEU scores of En$\to$Vi are reported in Table~\ref{tab:results_iwslt_tasks_eg}. For the \waitkstar, we also report the optimal training task (i.e., $k^*$) for each main task \waitk~. BLEU scores of all language pairs are left in Appendix D. Experiments on more heuristic baselines are in Appendix E.2.

\begin{table}[!htbp]
\centering
\small
\begin{tabular}{cccccc}
\toprule
$k$ & \waitk & \waitkstar / $k^*$ & CL & MTL & Ours \\
\midrule
$1$ & $25.14$ & $26.14$ / $5$ & $26.01$ & $26.12$ & $27.03$ \\
$3$ & $27.17$ & $28.25$ / $5$ & $26.37$ & $28.54$ & $29.01$ \\
$5$ & $28.29$ & $28.44$ / $9$ & $27.97$ & $28.61$ & $28.91$ \\
$7$ & $28.31$ & $28.38$ / $13$ & $28.31$ & $28.77$ & $29.17$ \\
$9$ & $28.39$ & $28.39$ / $9$ & $28.31$ & $28.70$ & $29.06$ \\
\bottomrule
\end{tabular}
\caption{BLEU scores on IWSLT En$\to$Vi task. 
}
\label{tab:results_iwslt_tasks_eg}
\end{table}

We have the following observations:

\noindent(1) Generally, our method consistently performed the best across different translation tasks in terms of both translation quality and controllable latency. 
As shown in Table~\ref{tab:results_iwslt_tasks_eg}, our method achieves the highest BLEU scores among all baselines. 
In Figure~\ref{fig:iwslt_bleu_latency}, the curve for our method (i.e., the red one) is on the top in most cases, which indicates that given specific latency (e.g., AP or AL), we can achieve the best translation quality. The results of significance tests are left in Appendix D.

\noindent(2) MTL and CL can outperform the vanilla baseline, which demonstrates the effectiveness of leveraging the temporally correlated tasks.
However, the improvements are not consistent, and it is hard to tell which baseline is better. CL slightly outperforms MTL on En$\to$De. However, on En$\to$Zh, CL performs better at lower latency while MTL performs better at higher latency. In comparison, the improvement brought by our method is much more consistent.

\noindent(3) \Waitkstar{} also outperforms \waitk{}. Specifically, we notice that the optimal training $k^*$ is always larger than $k$ for the main task. This is consistent with the observation in \citet{ma2019stacl}.
However, compared with CL and MTL, \waitkstar{} does not have a consistent advantage. This shows that the improvements of MTL, CL and our method is not due to the existence of some ``best'' task in $\domT$, and further demonstrates the importance of adaptively using all auxiliary tasks. 

\begin{figure}[!htbp]
\centering
\subfigure[BLEU-AP]{\includegraphics[width=0.5\linewidth]{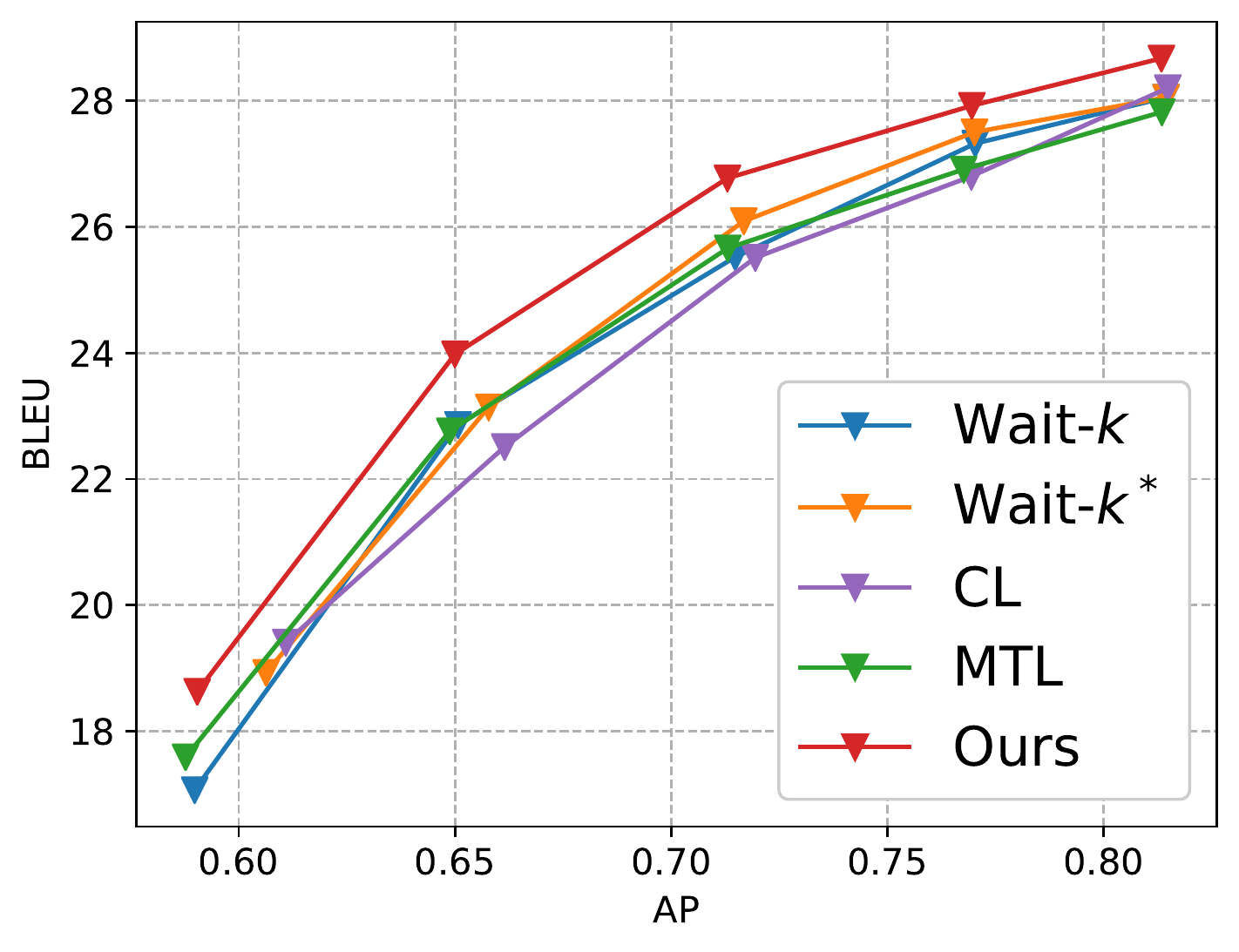}}%
\subfigure[BLEU-AL]{\includegraphics[width=0.5\linewidth]{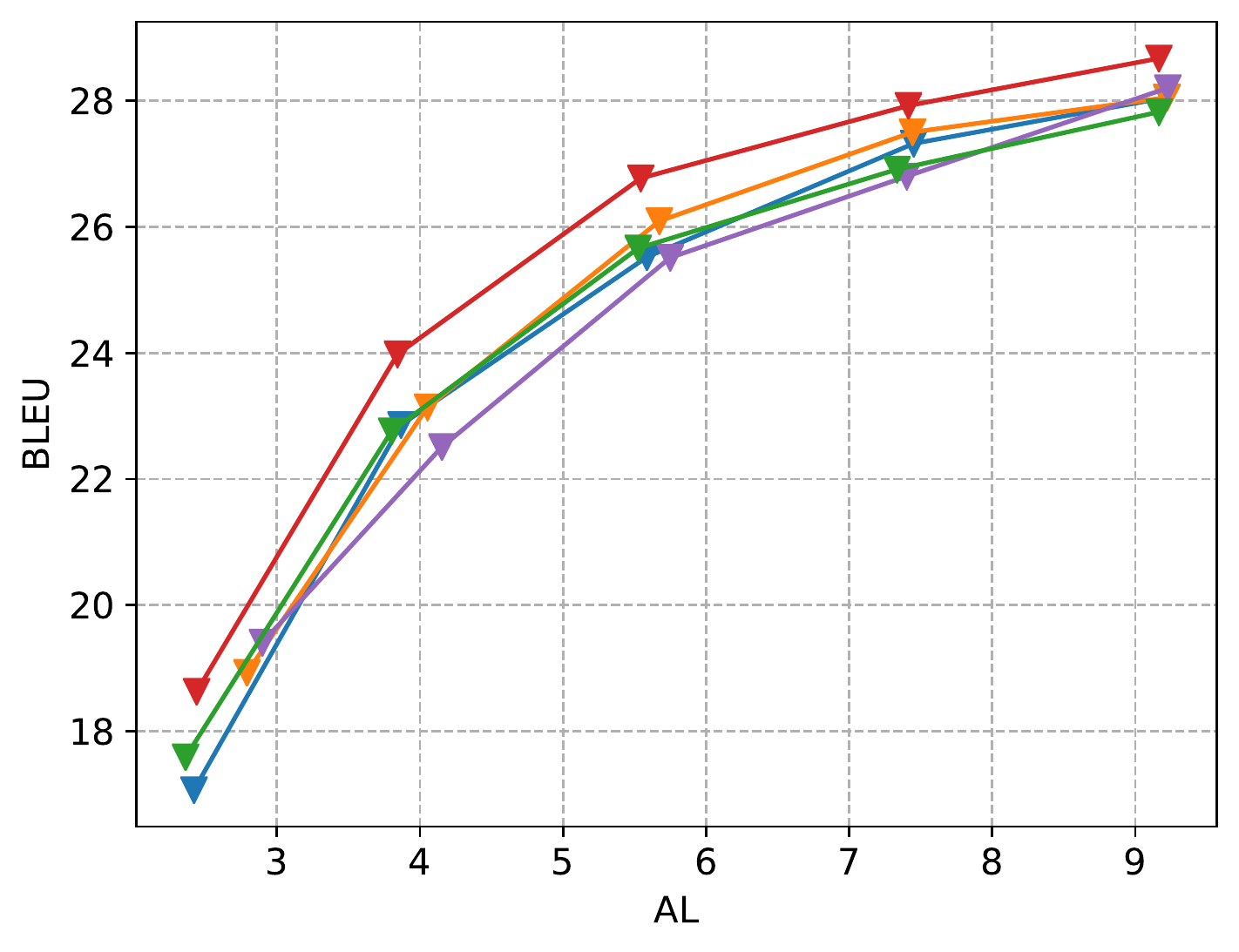}}
\caption{Translation quality against latency metrics (AP and AL) on WMT'15 En$\to$De.}
\label{fig:wmt15}
\end{figure}

The results on WMT'15 En$\to$De, whose training corpus is larger, are shown in Figure~\ref{fig:wmt15}. \Waitkstar{}, MTL and CL do not bring much improvement compared to vanilla \waitk{}. Our method consistently outperforms all baselines, which demonstrates the effectiveness of our method on large datasets. We further
evaluate them on WMT'14 and WMT'16 test sets and obtain similar conclusions (see Appendix D for details). 

\textit{Comparison with adaptive simultaneous NMT methods}:
We further compare our method with WIW, WID, MILk, MMA-IL and MMA-H on IWSLT En$\to$Vi.
The BLEU-AL curves are shown in Figure~\ref{fig:cmp_baseline_al} and the BLEU-AP curves are in Appendix D. When AL $\ge5.0$, our method outperforms all baseline models, and when AL $<5.0$, our method performs slightly worse than MMA-IL and MMA-H. A possible reason is that our method does not explicitly reduce the latency, but focuses on improving performance under given latency constraints. 

We further compare and combine our method with \citet{zheng2020simultaneous}.
We conduct experiments on two variants of \citet{zheng2020simultaneous}, where the \waitm{} models are obtained through vanilla \waitk{} (denoted by ``Zheng et al.'') and our strategy respectively (denoted by ``Zheng et al. + Ours''. The BLEU-AL curves are shown in Figure~\ref{fig:cmp_fix_to_adaptive_al}, and the BLEU-AP curves are in Appendix D.
We can see that: (1) our method catches up with \citet{zheng2020simultaneous}, which is built upon $10$ models in total ({\it wait-}$1,2,\cdots,10$); (2) after combing our approach with \citet{zheng2020simultaneous}, the performance can be further boosted, which shows that our method is complementary to adaptive inference strategies like  \citet{zheng2020simultaneous}.

\begin{figure}[!htbp]
\centering
\subfigure[Comparison with several adaptive methods.]{\includegraphics[width=0.48\linewidth]{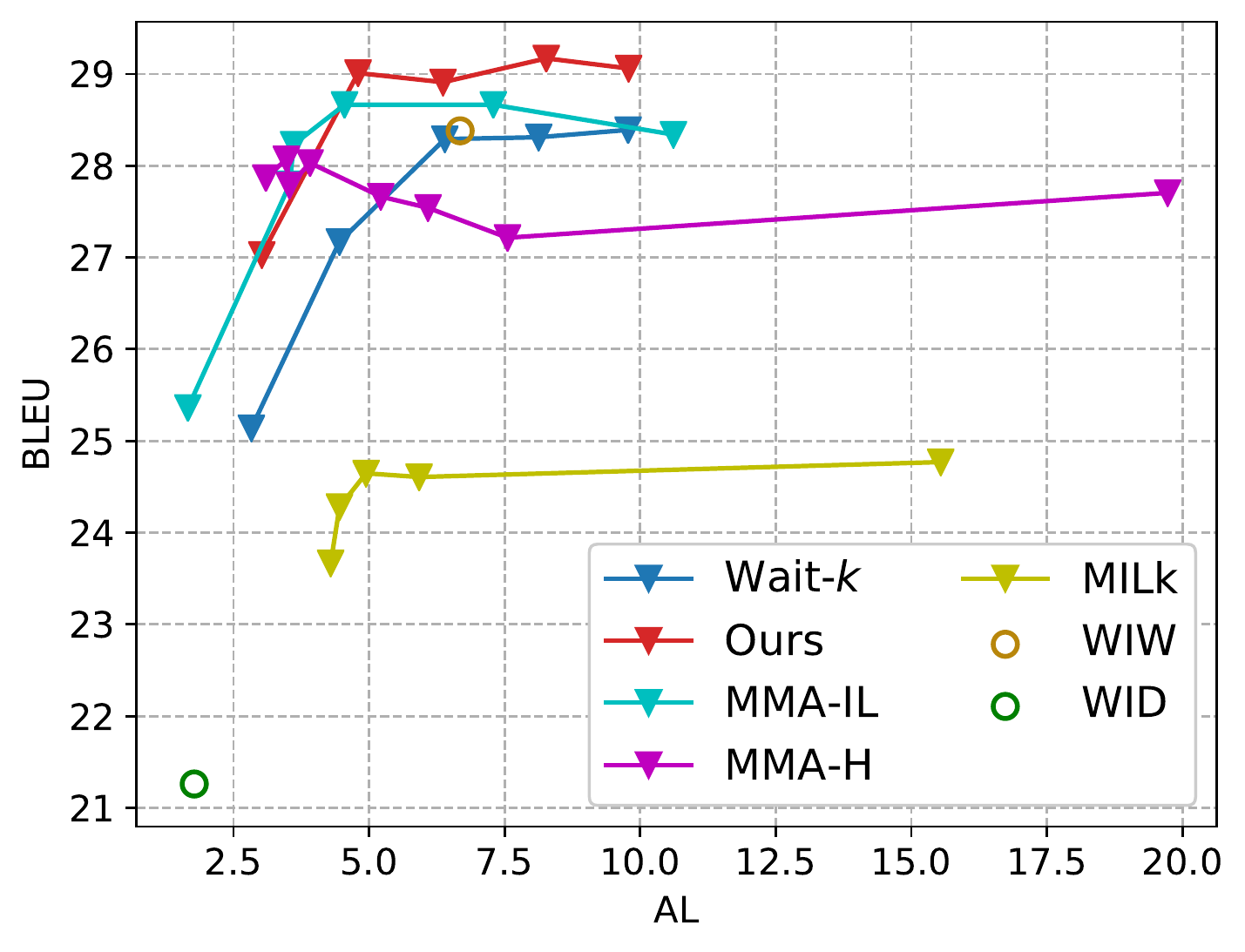}\label{fig:cmp_baseline_al}}%
\hspace{0.02\linewidth}
\subfigure[Comparison and combination with \citet{zheng2020simultaneous}.]{\includegraphics[width=0.48\linewidth]{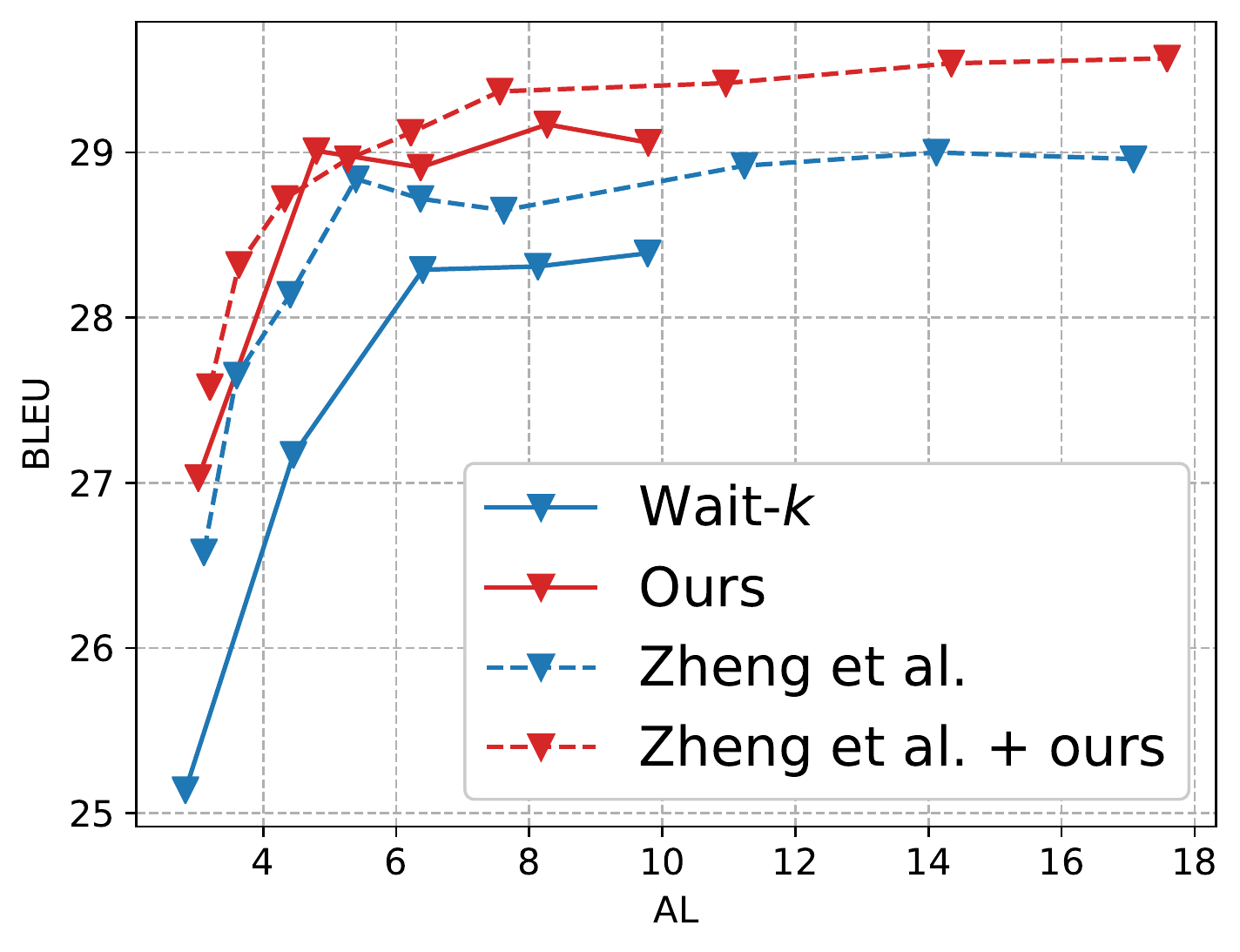}\label{fig:cmp_fix_to_adaptive_al}}
\caption{BLEU-AL comparison between our method and baselines on En$\to$Vi.}
\end{figure}

\textit{Computational overhead}: To evaluate the additional computational overhead brought by our method, we compare the training speed of standard \waitk{} and our method on {\it wait}-$3$ (measured by the number of batches per second). Results on IWSLT datasets are summarized in Table~\ref{tab:training_speed}.
Our method requires $20\%\sim30\%$ additional training time, which is acceptable considering the performance improvements. The major overhead comes from computing the training loss by \waitk{}. To verify that, we record the training speed of our method without the second and third input features of $\varphi$, i,e, the training loss over data $(x,y)$ evaluated on \waitk{}, and the average of historical training losses (denoted by ``Ours w/o feature (2,3)''). Without these two features, the training speed of our method is close to \waitk{}. 

\begin{table}[!htbp]
\small
\centering
\begin{tabular}{c|ccc}
\toprule
Task & \waitk & Ours & Ours w/o feature (2,3) \\
\midrule
En$\to$De & 5.3 & 4.0 (-23\%) & 5.2 (-2\%) \\
En$\to$Vi & 1.5 & 1.1 (-27\%) & 1.4 (-7\%) \\
En$\to$Zh & 2.5 & 1.8 (-28\%) & 2.4 (-4\%) \\
\bottomrule
\end{tabular}
\caption{Comparison of training speed (batch / sec) between \waitk{} and our methods.}
\label{tab:training_speed}
\end{table}

\begin{figure*}[!htpb]
\centering
\includegraphics[width=0.85\linewidth]{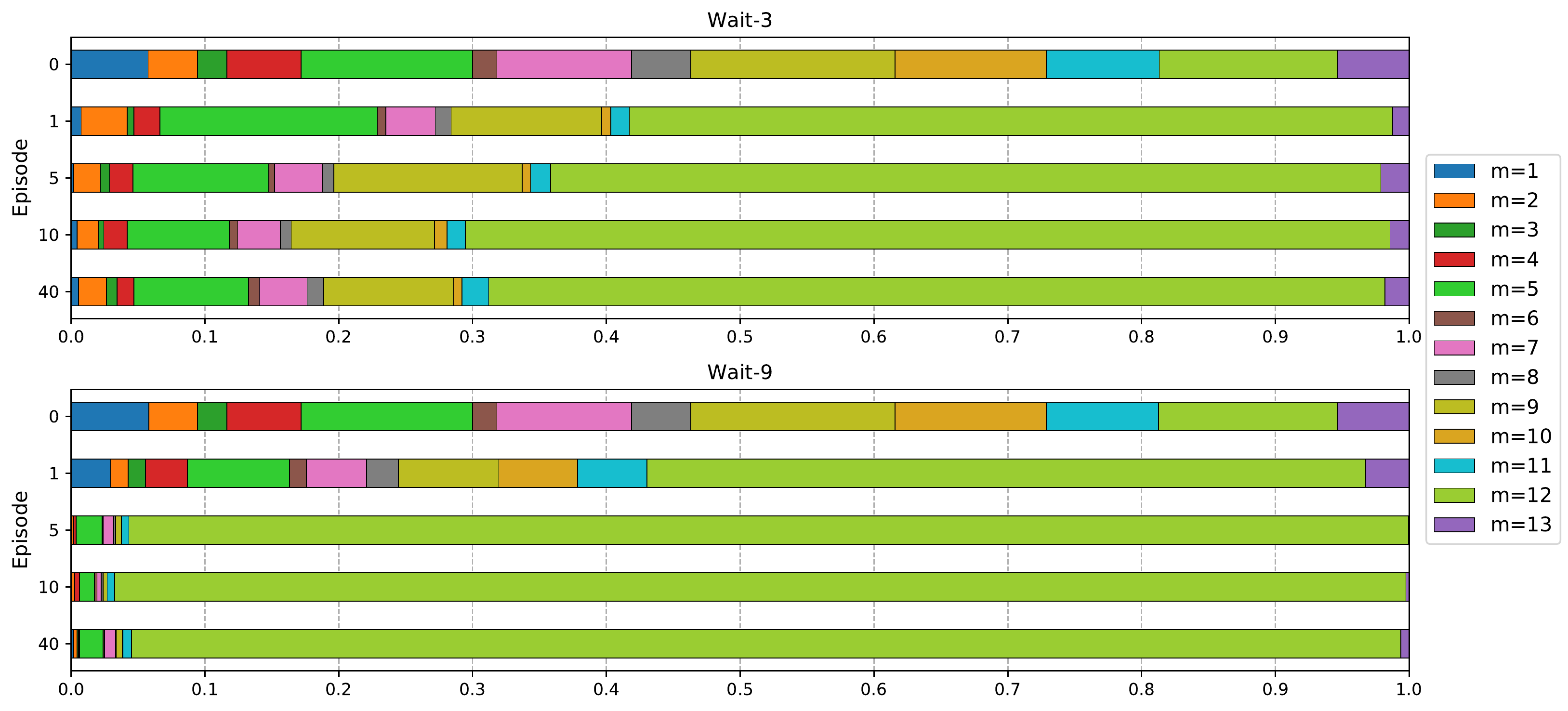}
\caption{The distribution of \waitm{} tasks for \textit{wait}-$3$ and \textit{wait}-$9$ on En$\to$Zh dataset.}
\label{fig:wait39_strategy}
\end{figure*}

\textit{Strategy analysis}:
In Figure~\ref{fig:wait39_strategy}, we visualize the distribution of \waitm{} tasks obtained by the scheduler for two main tasks, \textit{wait}-$3$ and \textit{wait}-$9$ on IWSLT En$\to$Zh dataset.
We show the frequency of each \waitm{} task sampled by the scheduler $\varphi$ at the $0$th, $1$st, $5$th, $10$th and $40$th episode. 

We observed that the scheduler uniformly samples different tasks at first, and then the strategies converge within $10$ episodes. After convergence, the scheduler mainly samples several specific tasks, i.e., $m=5,9,12$ for \textit{wait}-$3$, and $m=5,12$ for \textit{wait}-$9$. The task that both schedulers prefer most is $m=12$, which is close to the optimal training task \waitkstar{} ($k^*=11$) for both \textit{wait}-$3$ and \textit{wait}-$9$.
This demonstrates that the scheduler can learn which task is more beneficial to the main task.
However, it is worth noting that the scheduler also samples other tasks, which shows the importance of training with multiple tasks instead of one single optimal task.
For example, the scheduler for \textit{wait}-$9$ samples both \textit{wait}-$5$ and $9$ with a probability of about $0.1$.

\section{Application to Stock Trend Forecasting}\label{sec:stock}

\subsection{Task Description}\label{sec:task_def_stock}
Stock trend forecasting is a core problem in stock market investment. Following the common practice, in this task, $\domX$ represents the historical transactions data including both prices and trading volumes, and $\domY$ denotes the stock price movement in future days.

At the $t$-th day, a specific stock can be represented by a $5$-dimension feature vector $x_t$ including the opening price, closing price (denoted as $p_t$), the highest price, the lowest price and the trading volume. The task $T_k\in\domT$ is to forecast the stock price rise percent of the next $k$-th day, i.e., $y_t=\frac{p_{t+k}}{p_{t+k-1}}-1$. This task lies in the second type of classes as introduced in Section~\ref{sec:problem:Setup}. In stock forecasting, we usually make predictions based on a window of historical inputs of length $L$, i.e., $(x_{t-L+1},x_{t-L+2},\cdots,x_t)$. 
We explore three different main tasks in this work: forecasting the stock rise trend of the next $1$st, $2$nd and $3$rd days (i.e., $T_1$, $T_2$ and $T_3$). As mentioned by~\citet{tang2020multistep}, all the above three tasks are important because investors tend to invest in stocks with rising price trends in a continuous period, rather than caring the next trading day only.



\subsection{Background}
Traditional methods for stock trend forecasting are based on time-series analysis models, such as the Auto-regressive Integrated Moving Average (ARMA) and Kalman Filters~\cite{yan2015application}. With the development of deep learning, applying neural network to stock forecasting becomes a new trend due to its better performance on noisy data. GRU~\citep{chung2014empirical} is one of the most representative deep neural networks in stock trend forecasting~\citep{hu2018listening}. By utilizing the reset gate and update gate, GRU can adaptively choose the information to be kept or dropped, which makes it relatively robust to the diverse and noisy sequential data in the stock market. Other types of recurrent neural networks are also leveraged such as LSTM, SFM~\citep{zhang2017stock}, ALSTM~\citep{qin2017dual}, etc.
Recent works start to use graph neural networks (e.g. GAT~\cite{velivckovic2017graph}) to model different stocks so that the information can be propagated along with the pre-defined relations~\citep{feng2019temporal,kim2019hats}. 

In addition to neural networks, many machine learning methods have been successfully applied to forecast stock future trends.
A particularly powerful branch is the tree-based approach, such as LightGBM~\citep{NIPS2017_6449f44a} and Random Forest. They have outstanding performance and certain interpretability, so they are widely used in stock market investment~\citep{khaidem2016predicting,tan2019stock}.

\subsection{Settings}
We conduct experiments on individual stocks in the Chinese A-share market. We explore three groups of temporally correlated tasks: $\domT_{3}=\{T_1,T_2,T_3\}$, $\domT_{5}=\{T_1,T_2,\cdots,T_5\}$ and $\domT_{10}=\{T_1,T_2,\cdots,T_{10}\}$.

\noindent{\em Dataset}~\label{stock_dataset}:
We use the historical transaction data for $300$ stocks on CSI$300$~\cite{csi300} from 01/01/2008 to 08/01/2020. We split the data into training ($01/01/2008$-$12/31/2013$), validation ($01/01/2014$-$12/31/2015$), and test sets ($01/01/2016$-$08/01/2020$) based on the transaction time. All the data is provided by~\citet{yang2020qlib}.

\noindent{\em Models}:
The forecasting model $f$ is built on top of a $2$-layer GRU with hidden dimension $32$. The length of the input sequence (i.e., the $L$ in Section~\ref{sec:task_def_stock}) is $60$. The scheduler $\varphi$ is a single-layer feed-forward network with ReLU activation. The hidden size of $\varphi$ is $32$.

\noindent{\em Input features of $\varphi$}: The features consist of four parts: 
(1) the data $(x,y)\in\Dtrain$ itself;
(2) the training loss of the previous iteration; 
(3) the forecasting results of the previous iteration; 
(4) the labels of each task in $\domT$. 

\iftrue

\noindent{\em Baselines}: Similar to Section~\ref{sec:experiments_simmt:settings}, we implement MTL and CL as baselines by using temporally correlated tasks. In CL, we start from the easiest task $T_1$, and then gradually move to the most difficult one. Besides, we implement different models for stock forecasting for comparison, including  simple feed-forward networks (denoted by MLP), lightGBM (briefly, LGB)~\cite{NIPS2017_6449f44a} and GAT~\cite{velivckovic2017graph, feng2019temporal}. Note that for GAT, we do not leverage the specific relation between stocks.

\else

\noindent{\em Baselines}: As discussed in Section~\ref{sec:alg_part:discussion}, we implement MTL and CL as baselines by using temporally correlated tasks. In CL, we start from the easiest task $T_1$, and then gradually move to the most difficult one. Besides, we implement different models for stock forecasting for comparison, including simple feed-forward networks, lightGBM (briefly, LGB)~\cite{NIPS2017_6449f44a} and GAT~\cite{velivckovic2017graph, feng2019temporal}. Note that for GAT, we do not leverage the specific relation between stocks.

\fi

\noindent{\em Evaluation}:
We use rank correlation coefficient~\cite{zhang2020information}
(briefly, RankIC) and the mean square error (briefly, MSE) to evaluate the forecasting results. RankIC can be regarded as a correlation between the prediction and the groundtruth results (the larger, the better). MSE measures the gap between prediction results and the groundtruth results (the smaller, the better). 
Due to space limitation, we put the evaluation results using ICIR (a measurement that takes stability into consideration) in Appendix F.

\begin{table}[!htb]
\centering
\small
\begin{tabular}{lcccccc}
\toprule
Methods    & Task-1  & Task-2  & Task-3 \\
\midrule
GRU & 0.049 / 1.903 & 0.018 / 1.972 & 0.014 / 1.989 \\
MLP & 0.023 / 1.961 & 0.022 / 1.962 & 0.015 / 1.978 \\
LGB & 0.038 / 1.883 & 0.023 / 1.952 & 0.007 / 1.987 \\
GAT & 0.052 / 1.898 & 0.024 / 1.954 & 0.015 / 1.973 \\
\midrule
MTL ($\domT_3$) & 0.061 / 1.862 & 0.023 / 1.942 & 0.012 / \textbf{1.956} \\
CL ($\domT_3$) & 0.051 / 1.880 & 0.028 / 1.941 & 0.016 / 1.962 \\
Ours ($\domT_3$) & \textbf{0.071} / \textbf{1.851} & \textbf{0.030} / \textbf{1.939} & \textbf{0.017} / 1.963 \\
\midrule
MTL ($\domT_5$) & 0.057 / 1.875 & 0.021 / 1.939 & 0.017 / 1.959\\
CL ($\domT_5$) & 0.056 / 1.877 & 0.028 / 1.942 & 0.015 / 1.962 \\
Ours ($\domT_5$) & \textbf{0.075} / \textbf{1.849} & \textbf{0.032} /\textbf{1.939} & \textbf{0.021} / \textbf{1.955} \\
\midrule
MTL ($\domT_{10}$) & 0.052 / 1.882 & 0.020 / 1.947 & 0.019 / 1.952\\
CL ($\domT_{10}$) & 0.051 / 1.882 & 0.028 / 1.950 & 0.016 / 1.961 \\
Ours ($\domT_{10}$) & \textbf{0.067} /  \textbf{1.867} & \textbf{0.030} / \textbf{1.960} & \textbf{0.022} / \textbf{1.942} \\
\bottomrule
\end{tabular}
\caption{\small{Results of stock trend forecasting. The first four rows are the results of various model architectures without leveraging temporally correlated tasks. The following rows are the results with different algorithms (MTL, CL and ours) and different auxiliary task sets ($\domT_3$, $\domT_5$ and $\domT_{10}$). For each cell, the RankIC is put on the left and MSE on the right.}}
\label{tab:results_of_stock_forecasting}
\end{table}

\subsection{Results and Analysis}
The results of stock trend forecasting are reported in Table~\ref{tab:results_of_stock_forecasting}. We have the following observations:

\noindent(1) With different sets of temporally correlated auxiliary tasks, i.e., $\domT_{3}$, $\domT_{5}$ and $\domT_{10}$, our method significantly outperforms MTL and CL on most settings, which demonstrates the effectiveness of leveraging an adaptive scheduler.

\noindent(2) Using a larger temporally correlated task set does not always bring improvements.
Taking task $T_1$ as an example, our method achieves the best performance when using $5$ temporally correlated tasks, while using $10$ hurts the performance. The reason is that more tasks will introduce additional noise to the training.

\noindent(3) There are more advanced methods than GRU like GAT, leveraging temporally correlated tasks can achieve superior performance than the strong models. Currently, our model is based on GRU, and we will combine our method with GAT and lightGBM in the future for more improvements.

\section{Conclusions and Future Work}\label{sec:conclusion}
In this work, we formulated and studied the problem of temporally correlated task scheduling, which can cover many applications including sequence processing with latency constraints (e.g., simultaneous machine translation) and sequence future prediction (e.g. stock trend forecasting). Experiments on four translation tasks and one stock prediction task demonstrate the effectiveness of our approach.

For future work, we will apply our method to more applications like action prediction~\citep{8543243,cai2019action}, streaming speech recognition, weather forecasting, game AI~\citep{li2020suphx,vinyals2019grandmaster}, etc. We will also design better methods for the problem. Another interesting direction is to theoretically understand the problem and analyze our method.

\bibliography{myref}
\bibliographystyle{icml2021}

\clearpage
\onecolumn
\appendix

\section{Mathematical definitions}

\subsection{Latency metrics definitions}\label{appendix:apal}

Given the input sentence $x$ and the output sentence $y$, let $L_x$ and $L_y$ denote the length of $x$ and $y$ respectively.
Define a function $g(t)$ of decoding step $t$,
which denotes the number of source tokens processed by the encoder when deciding the target token $y_t$.
For \waitk{} strategy, $g(t) = \min\{t + k - 1, L_x\}$. The definitions of Average Proportion (AP) and Average Lagging (AL) are  in Eqn.(\ref{eq:ap}) and Eqn.(\ref{eq:al}) respectively.
\begin{align}
& \textmd{AP}_g(x, y) = \frac{1}{|x||y|} \sum_{t=1}^{|y|} g(t); \label{eq:ap} \\
& \textmd{AL}_g(x, y) = \frac{1}{\tau_g(|x|)}  \sum_{t=1}^{\tau_g(|x|)} \left( g(t) - \frac{t-1}{|y| / |x|} \right), \label{eq:al} \\
& \qquad \textmd{where} \quad \tau_g(|x|) = \min\{ t | g(t) = |x|\}. \notag
\end{align}
We use the scripts provided by \citet{ma2019stacl} to calculate AP and AL scores.

\subsection{Mathematical formulation of curriculum transfer learning} \label{appendix:cl}

For a group of temporally correlated tasks $\domT=\{T_1,T_2,\cdots,T_M\}$, suppose $\domT$ is ordered by task difficulty. That is, $T_M$ is the easiest task, followed by $T_{M-1}$, $T_{M-2}$, etc, and $T_1$ is the hardest task.
In the curriculum transfer learning (briefly, CL) baseline, we gradually change the training task $T_m$ from the easiest task $T_M$ to the main task, supposedly, $T_k$.
The mathematical formulations are shown as follows:
\begin{align}
m = M - \lfloor \frac{t - 1}{t_\text{max}} (M - k + 1) \rfloor 
\end{align}
where $t$ denotes the current update number ($t = 1, 2, ..., t_\text{max}$), and $t_\text{max}$ denotes the total update number. 

\section{Model architecture}\label{appendix:ablation_arch}

Following \citet{ma2019stacl}, our model for simultaneous NMT is based on Transformer model~\citep{vaswani2017attention}. The model includes an encoder and a decoder, which are used for incrementally processing the source and target sentences respectively. Both the encoder and decoder are stacked of $L$ blocks.

In \citet{ma2019stacl}, every time a new token is read, the encoder needs to re-encode the entire sentence, which is computationally expensive. To improve the efficiency, we adopt incremental encoding, so that after reading a new token, we only encode the new token rather than updating the representations of the whole sentence. The details are as follows:


\noindent(1) {\em Incremental encoding}: Let $h^l_t$ denote the output of the $t$-th position from block $l$. 
For ease of reference, let $H_{\le t}^l$ denote $\{h^l_1,h^l_2,\cdots,h^l_{t}\}$, and let $h^0_t$ denote the embeddings of the $t$-th token. An attention model $\attn(q,K,V)$, takes a query $q\in\mathbb{R}^d$, a set of keys $K$ and values $V$ as inputs. $K$ and $V$ are of equal size, $q\in\mathbb{R}^d$ where $d\in\mathbb{Z}_+$ is the dimension, $k_i\in\mathbb{R}^d$ and $v_i\in\mathbb{R}^d$ are the $i$-th key and value. $\attn$ is defined as follows:
\begin{equation}
\attn(q,K,V)=\sum_{i=1}^{\vert K\vert}\alpha_iW_vv_i,\;\alpha_i=\frac{\exp((W_qq)^\top(W_kk_i))}{Z},\;Z=\sum_{i=1}^{\vert K\vert}\exp((W_qq)^\top(W_kk_i)),
\end{equation}
where $W$'s are the parameters to be optimized. In the encoder side, $h_t^l$ is obtained in a unidirectional way: $h^{l}_{t} = \attn(h^{l-1}_{t},H_{\le t}^{l-1},H_{\le t}^{l-1}).$
That is, the model can only attend to the previously generated hidden representations, and the computation complexity is $O(L^2_x)$. In comparison, \citep{ma2019stacl} still leverages bidirectional attention, whose computation complexity is $O(L_x^3)$. 
We find that unidirectional attention is much more efficient than bidirectional attention without much accuracy drop (see Appendix \ref{appendix:ablation_arch} for details). 

\noindent(2) {\it Incremental decoding}: Since we use \waitk{} strategy, the decoding starts before reading all inputs.
At the $t$-th decoding step, the decoder can only read $x_{\le t+k-1}$.
When $t \le L_x - k$, the decoder greedily generates one token at each step, i.e., the token is $y_{t}=\argmax_{w\in\mathcal{V}} P(w|y_{\le t-1};H^L_{\le t + k - 1})$, where $\mathcal{V}$ is the vocabulary of the target language. When $t > L_x - k$, the model has read the full input sentence and can generate words using beam search~\citep{ma2019stacl}.

\begin{figure}[!htbp]
	\centering
	\subfigure[BLEU-AP, IWSLT'14 En$\to$De]{
		\includegraphics[width=0.4\linewidth]{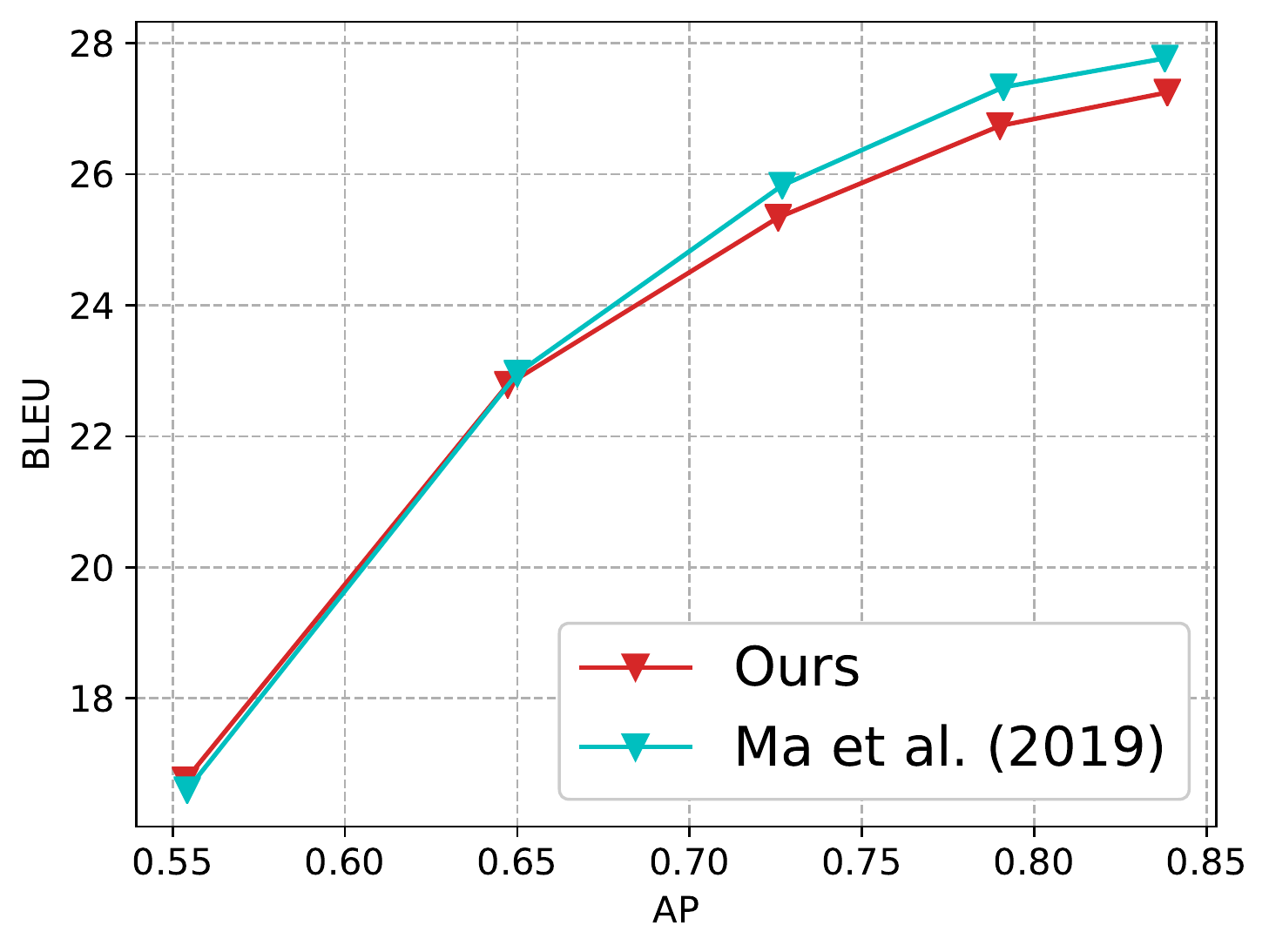}
		\label{fig:ablation_arch_ap}
	}
	\subfigure[BLEU-AL, IWSLT'14 En$\to$De]{
		\includegraphics[width=0.4\linewidth]{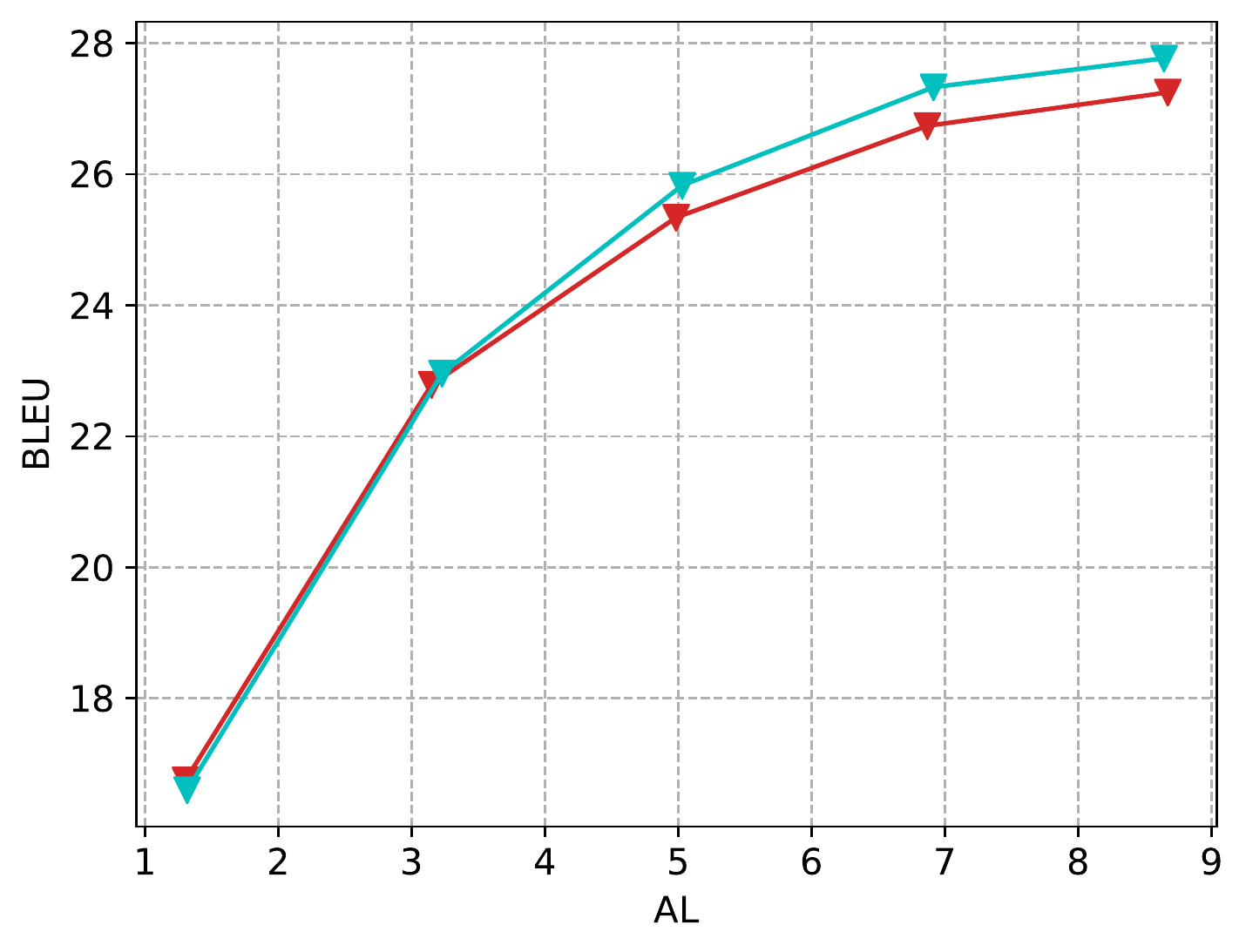}
		\label{fig:ablation_arch_al}
	}
	\caption{Ablation study of different model architectures on IWSLT'14 En$\to$De dataset and WMT'15 En$\to$De dataset.}
	\label{fig:ablation_arch}
\end{figure}

We compare the performance of original \citet{ma2019stacl} and our modified model on IWSLT'14 En$\to$De dataset, and the results are in Figure~\ref{fig:ablation_arch},
On IWSLT'14, we observe that the performance of our model slightly drops compared to \citet{ma2019stacl}.
However, the computational cost of \citet{ma2019stacl} is much larger than our modified model.
For example, the inference speed of our \textit{wait}-$9$ model is $57.39$ sentences / second, while the inference speed of \citet{ma2019stacl} is $6.48$ sentences / second.

\section{More detailed settings about experiments}
\subsection{Detailed introduction of the datasets}\label{appendix:dataset}
For IWSLT'14 En$\to$De, following~\cite{edunov-etal-2018-classical}, 
we lowercase all words, tokenize them and apply BPE with $10k$ merge operations jointly to the source and target sequences. We split $7k$ sentences from the training corpus for validation and the remaining $160k$ sequences are left as the training set. The test set is the concatenation of {\em tst2010, tst2011, tst2012, dev2010} and {\em dev2012}, which consists of $6750$ sentences.

For IWSLT'15 En$\to$Vi, following~\cite{ma2020monotonic}, 
we tokenize the data and replace words with frequency less than 5 by \texttt{<unk>}\footnote{The data is downloaded from \url{https://nlp.stanford.edu/projects/nmt/}, which has been tokenized already.}. We use {\it tst2012} as the validation set and {\it tst2013} as the test set. The training, validation and test sets contains $133k$, $1268$ and $1553$ sentences respectively.

For IWSLT'17 En$\to$Zh, we tokenize the data and apply BPE with $10k$ merge operations independently to the source and target sequences\footnote{The Chinese sentences are tokenized using Jieba ( \url{https://github.com/fxsjy/jieba} ).}. We use the concatenation of {\it tst2013}, {\it tst2014} and {\it tst2015} as the validation set and use {\it tst2017} as the test set. The training, validation and test sets contains $235k$, $3874$ and $1459$ sentences respectively.
For WMT'15 En$\leftrightarrow$De, we follow the setting in~\cite{ma2019stacl, arivazhagan-etal-2019-monotonic}. 
We tokenize the data, apply BPE with $32k$ merge operations jointly to the source and target sentences, and get a training corpus with $4.5M$ sentences. We use {\it newstest2013} as the validation set and use {\it newstest2015} as the test set.

\subsection{Detailed training strategy}\label{sec:training_details}
For the translation model, we use Adam~\citep{adam_optimizer} optimizer with initial learning rate $5\times10^{-4}$ and \texttt{inverse\_sqrt} scheduler (see Section 5.3 of \citep{vaswani2017attention} for details). The batch size and the number of GPUs of IWSLT En$\to$De, En$\to$Vi, En$\to$Zh and WMT'15 En$\to$De are $4096\times1$GPU, $16000\times1$GPU, $4000\times1$GPU and $3584 \times 8 \times 16$GPU respectively. For 
IWSLT tasks, the learning rate $\eta$ is grid searched from $\{5\times10^{-4},5\times10^{-5},5\times10^{-6},5\times10^{-7}\}$ with vanilla SGD optimizer, and the internal update iteration $S$ is grid searched from $\{\frac{1}{2}
S_e, S_e, 2S_e\}$, where $S_e$ is the number of updates in an epoch of the translation model training. 
For WMT'15 En$\to$De, due to resource limitation, we do not train the translation model from scratch. The translation model is warm started from pretrained \waitk{} model, the learning rate is set as $5\times10^{-5}$, and the internal update iteration $S$ is $16$.

\subsection{Detailed baseline implementation}\label{sec:impl_details}

In this section, we introduce how we reproduce baseline models and get the results on En$\to$Vi task.

For MILk \citep{arivazhagan-etal-2019-monotonic} and MMA \citep{ma2020monotonic}, we do not reproduce the results, but directly use the results reported in \citet{ma2020monotonic}. Note that the results for MILk is reproduced by \citet{ma2020monotonic}.

For WIW and WID \citep{DBLP:journals/corr/ChoE16}, we pre-train a standard translation model, using the exact same architecture and training hyperparameters as our method and in \citep{ma2020monotonic}. For fair comparison, we adopt bi-directional attention.

For \citet{zheng2020simultaneous}, we use the pre-trained \waitk{} model and the model obtained through our results ($k=1,2,3,...,10$). We use only single model rather than ensemble. As in \citet{zheng2020simultaneous}, the thresholds of different $k$ values are obtained in this way: $\rho_i = \rho_1 - (i - 1) * (\rho_1 - \rho_{10}) / 9$, where we test with $\rho_1\in \{0.2, 0.4, 0.6, 0.8\}$, $\rho_{10}=0$; and $\rho_{1}=1$, $\rho_{10} \in \{0, 0.2, 0.4, 0.6, 0.8\}$.

\section{Supplemental results}\label{appendix:supp_results}

\begin{figure}[!htbp]
	\centering
	\subfigure[Comparison of our method, WIW, WID, MILk, MMA-IL and MMA-H.]{\includegraphics[width=0.4\linewidth]{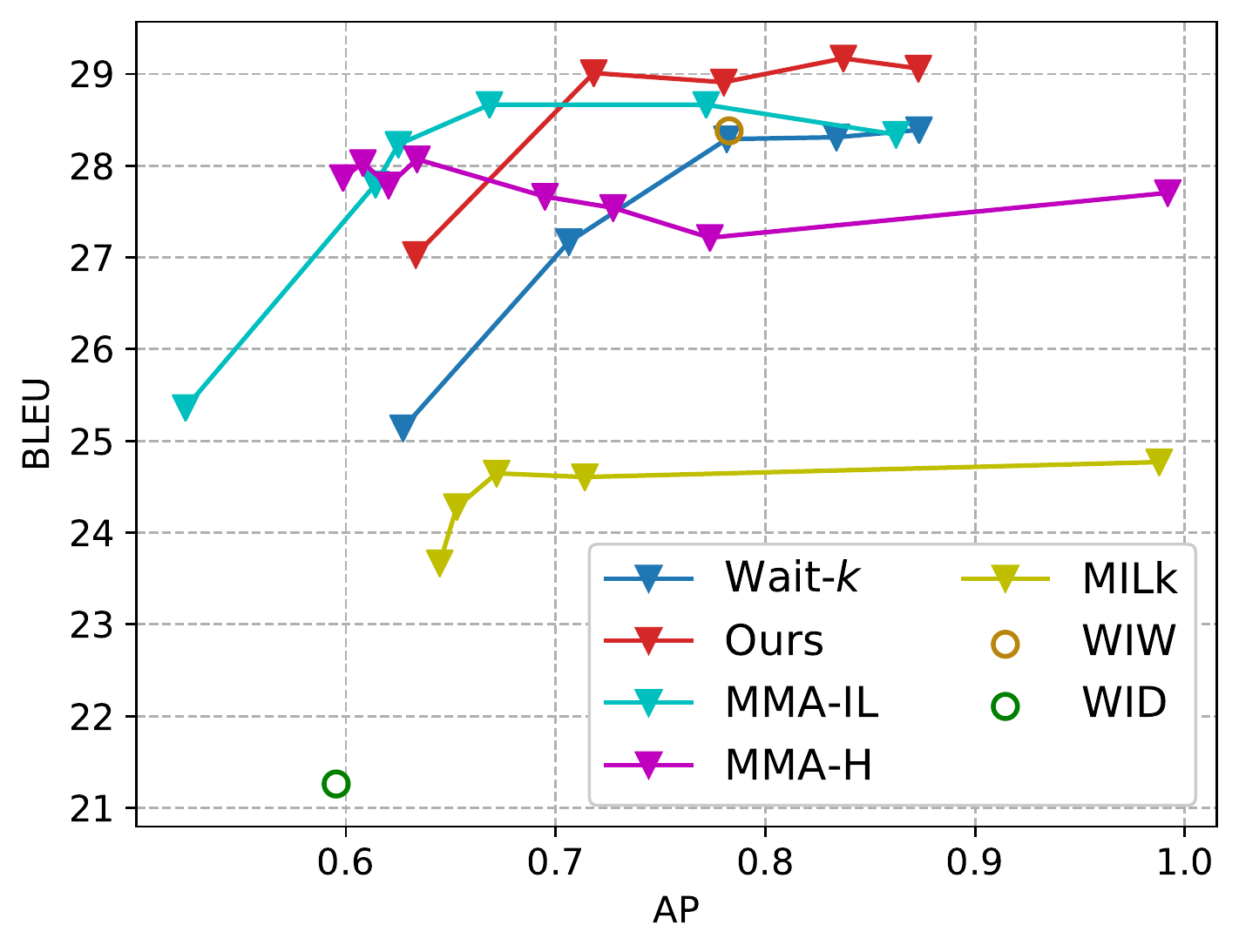}\label{fig:cmp_baseline_ap}}
	\hspace{0.05\textwidth}
	\subfigure[Comparison and combination of our method and \citet{zheng2020simultaneous}.]{\includegraphics[width=0.4\linewidth]{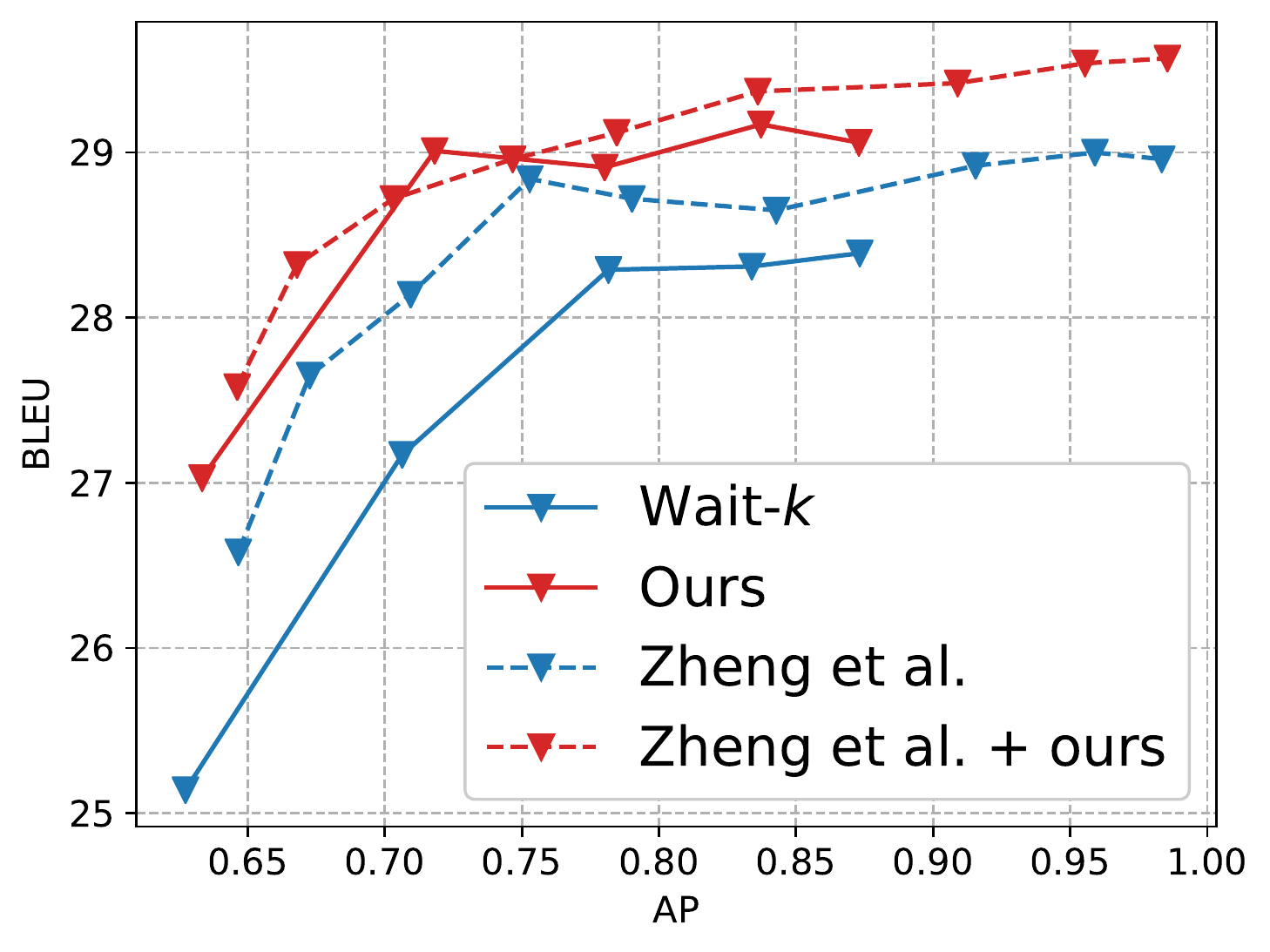}\label{fig:cmp_fix_to_adaptive_ap}}
	\caption{BLEU-AP comparison between our method and baselines on En$\to$Vi.}
\end{figure}

\begin{figure}[!htbp]
	\centering
	\subfigure[BLEU-AP, WMT'14 En$\to$De]{
		\includegraphics[width=0.4\linewidth]{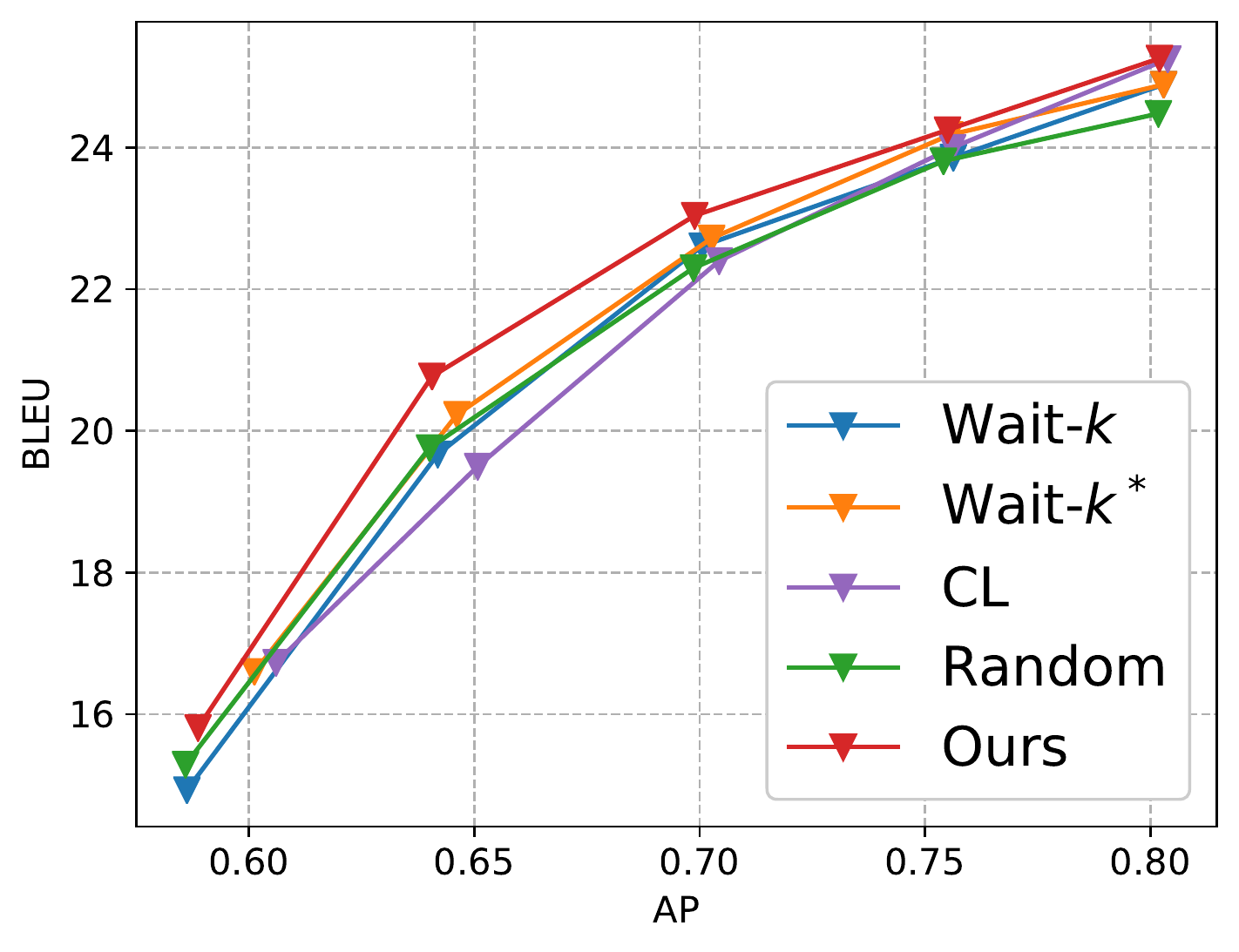}
	}
	\subfigure[BLEU-AP, WMT'16 En$\to$De]{
		\includegraphics[width=0.4\linewidth]{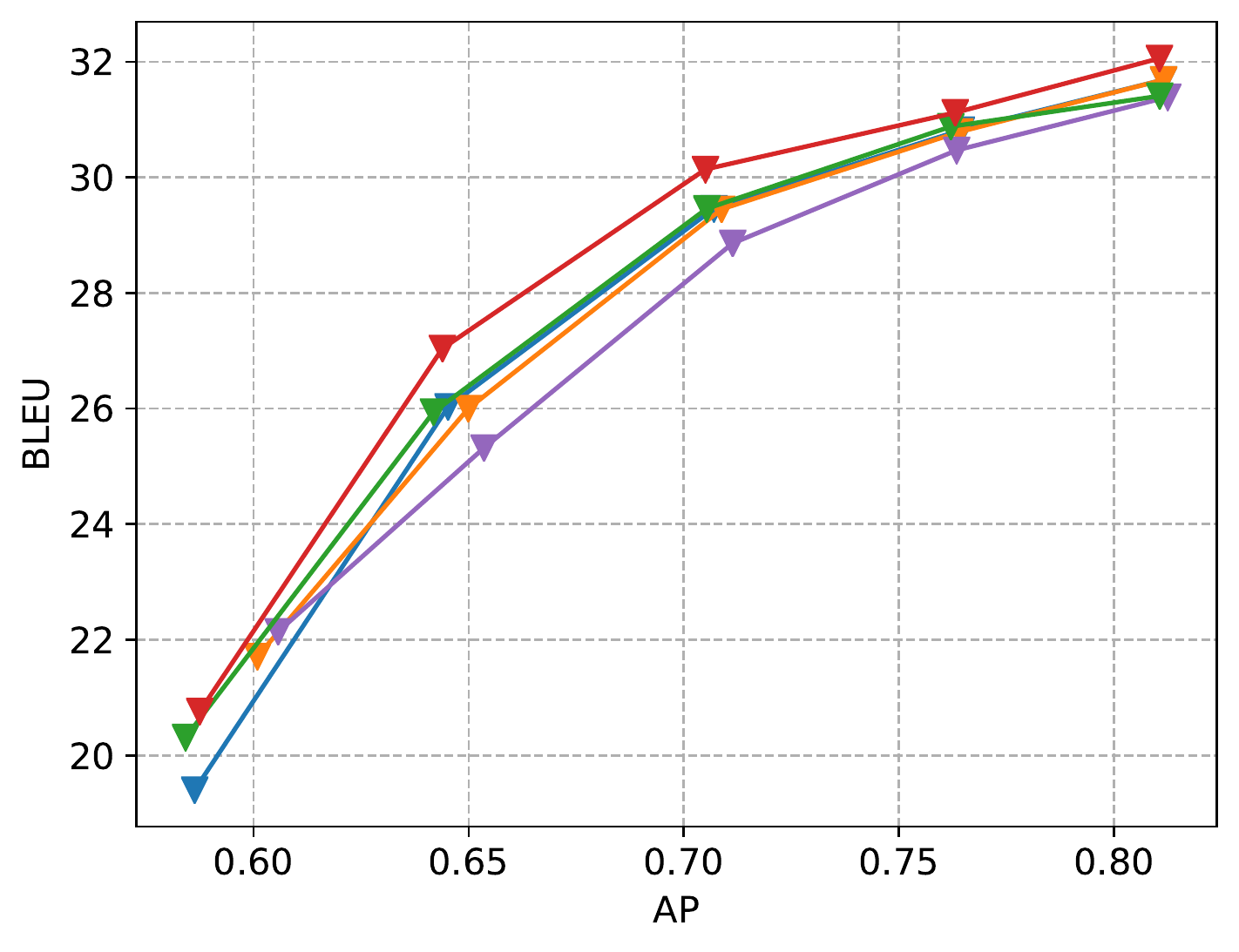}
	}
	
	\subfigure[BLEU-AL, WMT'14 En$\to$De]{
		\includegraphics[width=0.4\linewidth]{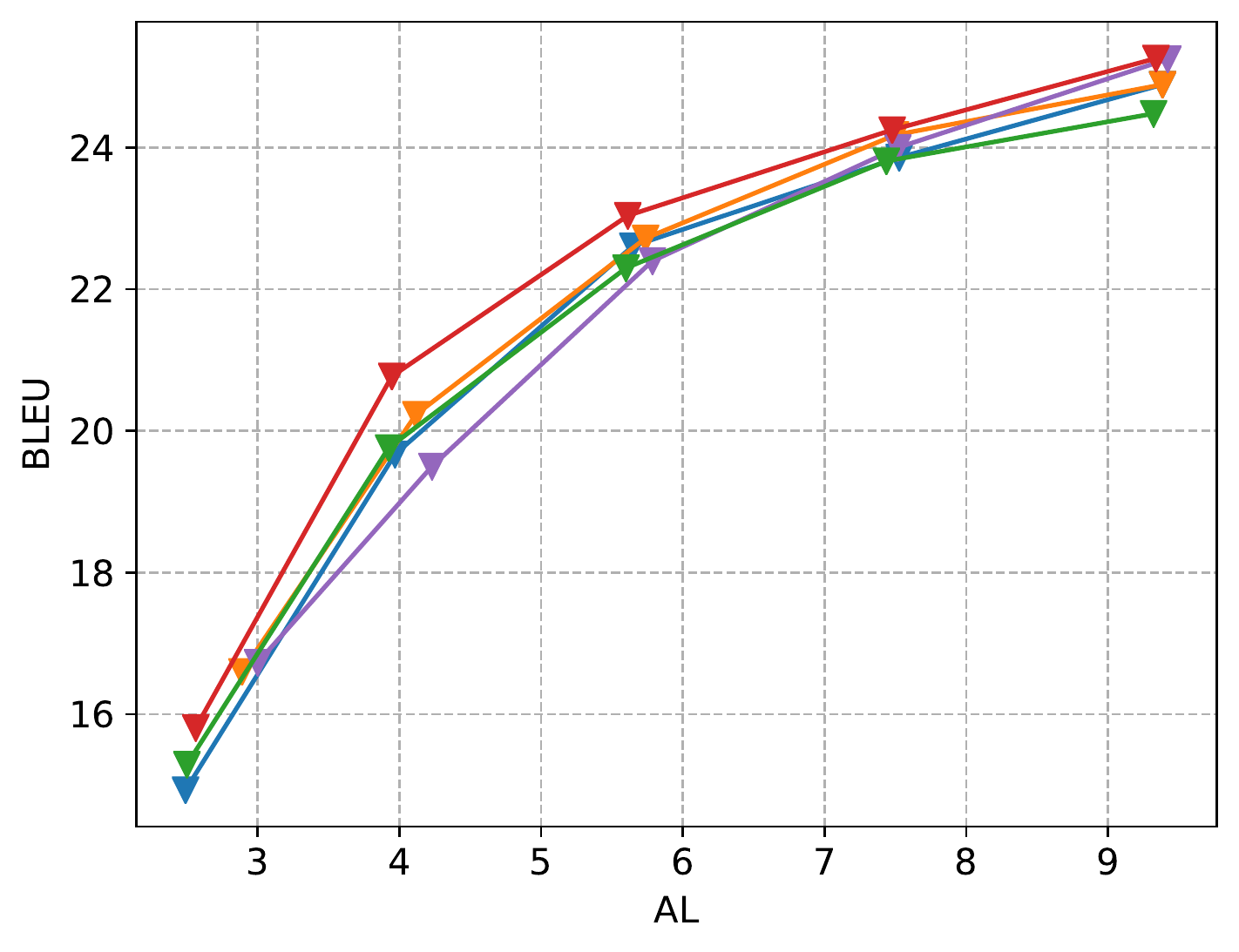}
	}
	\subfigure[BLEU-AL, WMT'16 En$\to$De]{
		\includegraphics[width=0.4\linewidth]{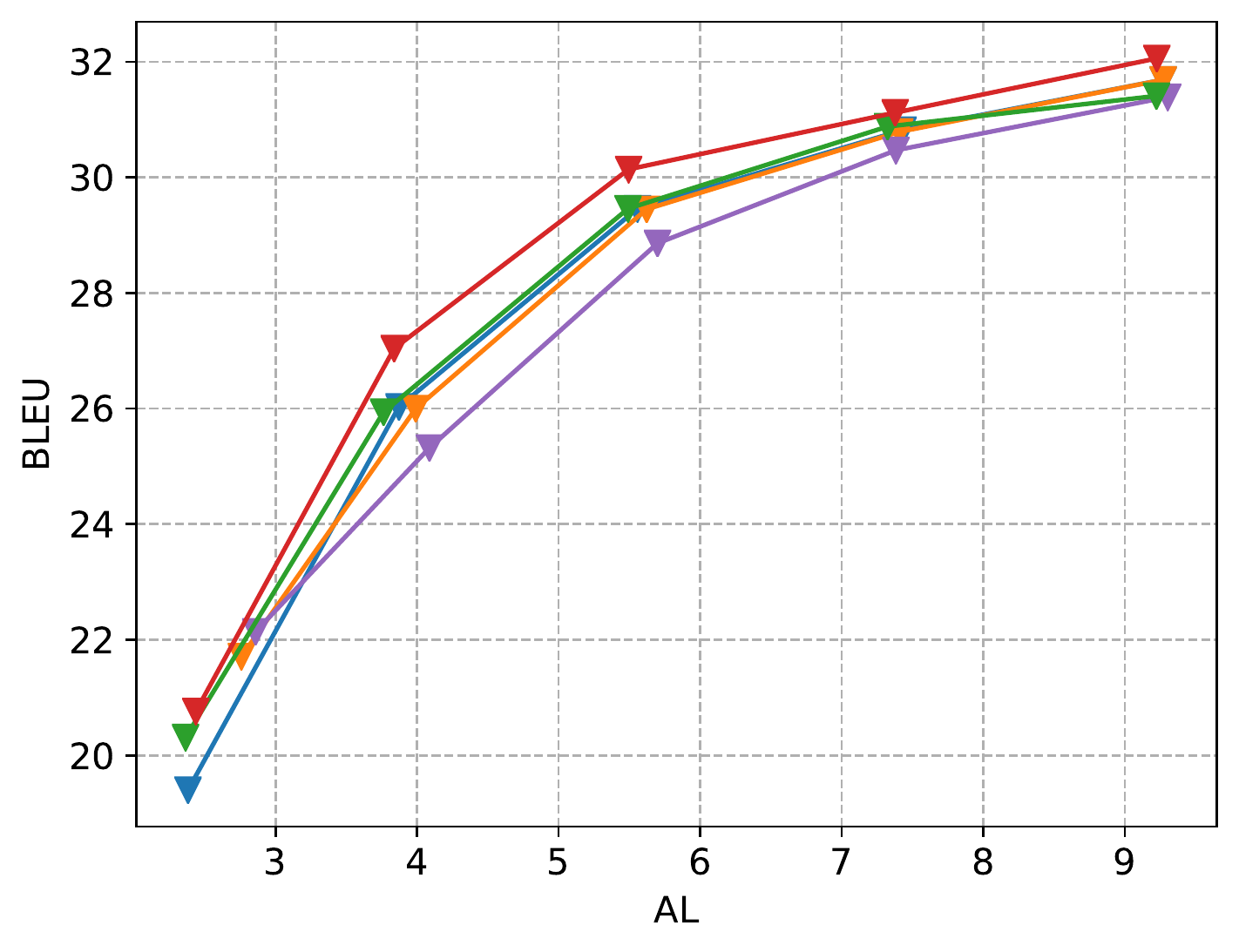}
	}
	\caption{Translation quality against latency metrics (AP and AL) on WMT'14 and 16 English$\to$German test sets.}
	\label{fig:wmt_manytestsets}
\end{figure}

In this section, we report the specific BLEU scores and some additional results. 
The BLEU scores for IWSLT tasks and WMT'15 En$\to$De task are reported in Table~\ref{tab:results_iwslt_tasks} and Table~\ref{tab:results_wmt} respectively.
We further evaluate the baselines and our methods on WMT'14 and WMT'18 test sets, and report the BLEU-latency curves in Figure~\ref{fig:wmt_manytestsets}.
The BLEU-AL curves of our methods and baselines on IWSLT'15 En$\to$Vi are reported in Figure~\ref{fig:cmp_baseline_ap}, the BLEU-AL curves of our method and \citet{zheng2020simultaneous} are in Figure~\ref{fig:cmp_fix_to_adaptive_ap}.

For significance tests, we  compare our method with MTL on all 20 translation tasks (En$\to$\{Vi, Zh\} tasks and two En$\to$De tasks, where each task is associated with $5$ settings, i.e. {\it wait}-1,3,5,7,9), and find that our method significantly outperforms MTL on $16$ tasks with $p<0.05$, where $11$ out of them are with $p<0.01$.

\begin{table}[!htbp]
	\centering
	\begin{tabular}{cccccccccc}
		\toprule
		Task & \waitk{} & \waitkstar / best $k^*$ & CL & MTL & Ours \\
		\midrule
		En$\to$De ($k=1$) & $16.75$ & $19.11$ / $9$ & $18.23$ & $18.53$ & $19.10$ \\
		En$\to$De ($k=3$) & $22.79$ & $23.36$ / $13$ & $23.41$ & $23.50$ & $24.12$ \\
		En$\to$De ($k=5$) & $25.34$ & $25.76$ / $11$ & $25.88$ & $25.84$ & $26.54$ \\
		En$\to$De ($k=7$) & $26.74$ & $26.87$ / $9$ & $26.85$ & $26.88$ & $27.37$ \\
		En$\to$De ($k=9$) & $27.25$ & $27.54$ / $11$ & $27.48$ & $27.07$ & $27.87$ \\
		\midrule
		En$\to$Vi ($k=1$) & $25.14$ & $26.14$ / $5$ & $26.01$ & $26.12$ & $27.03$ \\
		En$\to$Vi ($k=3$) & $27.17$ & $28.25$ / $5$ & $26.37$ & $28.54$ & $29.01$ \\
		En$\to$Vi ($k=5$) & $28.29$ & $28.44$ / $9$ & $27.97$ & $28.61$ & $28.91$ \\
		En$\to$Vi ($k=7$) & $28.31$ & $28.38$ / $13$ & $28.31$ & $28.77$ & $29.17$ \\
		En$\to$Vi ($k=9$) & $28.39$ & $28.39$ / $9$ & $28.31$ & $28.70$ & $29.06$ \\
		\midrule
		En$\to$Zh ($k=1$) & $14.24$ & $19.34$ / $9$ & $17.26$ & $18.02$ & $19.21$ \\
		En$\to$Zh ($k=3$) & $19.90$ & $21.66$ / $11$ & $21.64$ & $21.36$ & $22.18$ \\
		En$\to$Zh ($k=5$) & $21.45$ & $23.57$ / $11$ & $23.62$ & $22.59$ & $23.70$ \\
		En$\to$Zh ($k=7$) & $23.23$ & $24.95$ / $11$ & $24.32$ & $23.15$ & $24.35$ \\
		En$\to$Zh ($k=9$) & $23.93$ & $24.83$ / $11$ & $24.55$ & $23.55$ & $24.78$ \\
		\bottomrule
	\end{tabular}
	\caption{BLEU scores on IWSLT simultaneous NMT tasks.} 
	\label{tab:results_iwslt_tasks}
\end{table}

\begin{table}[!htbp]
	\centering
	\begin{tabular}{cccccccccc}
		\toprule
		$k$ & \waitk{} & \waitkstar / best $k^*$ & CL & MTL & Ours \\
		\midrule
		$1$ & $17.07$ & $19.83$ / $9$ & $19.41$ & $17.59$ & $18.63$ \\
		$3$ & $22.86$ & $23.14$ / $7$ & $22.51$ & $22.76$ & $23.98$ \\
		$5$ & $25.52$ & $26.09$ / $7$ & $25.51$ & $25.66$ & $26.77$ \\
		$7$ & $27.32$ & $27.50$ / $9$ & $26.80$ & $26.91$ & $27.92$ \\
		$9$ & $28.05$ & $28.05$ / $9$ & $28.20$ & $27.82$ & $28.67$ \\
		\bottomrule
	\end{tabular}
	\caption{BLEU scores on WMT'15 En$\to$De dataset.}
	\label{tab:results_wmt}
\end{table}

\section{Additional ablations and analysis on simultaneous translation}

\subsection{Feature selection}

To emphasize the importance of the selected features in Section 3.3 in the main content, we provide four groups of ablation study, where in each group some specific features are excluded: (i) source and target sentence lengths; (ii) current training loss and average historical training loss; (iii) current validation loss and average historical validation loss; (iv) training step. We work on IWSLT'14 En$\to$De task and study the effect to \textit{wait-}$3,5,7$.

The results are shown in Table~\ref{tab:ablation_feature}. We report the BLEU scores only, since the latency metrics (AP and AL) are not significantly influenced. Removing any feature causes performance drop, indicating that they all contribute to the decision making. Specifically, network status information including validation performance (feature iii) and training stage (feature iv) is more important than input data information including sequence length (feature i) and data difficulty (feature ii).

%

\begin{table}[!htbp]
	\centering
	\begin{tabular}{llll}
		\toprule
		& $k=3$ & $k=5$ & $k=7$ \\
		\midrule
		Ours & 24.12 & 26.54 & 27.37 \\
		- (i) & 23.67 (-1.87\%, rank 3) & 26.03 (-1.92\%, rank 3) & 26.92 (-1.64\%, rank 4) \\
		- (ii) & 23.70 (-1.74\%, rank 4) & 26.04 (-1.88\%, rank 4) & 26.91 (-1.68\%, rank 3) \\
		- (iii) & 23.57 (-2.28\%, rank 1) & 25.92 (-2.34\%, rank 2) & 26.72 (-2.37\%, rank 1) \\
		- (iv) & 23.65 (-1.95\%, rank 2) & 25.63 (-3.43\%, rank 1) & 26.86 (-1.86\%, rank 2) \\
		\bottomrule
	\end{tabular}
	\caption{Ablation study for feature selection on IWSLT'14 En$\to$De dataset. 
	}
	\label{tab:ablation_feature}
\end{table}

\subsection{More heuristic baselines}

In this section, we conduct ablation studies on two more heuristic baselines to demonstrate the effectiveness of our method.

\begin{figure}[!htbp]
	\centering
	\subfigure[BLEU-AP, IWSLT'14 En$\to$De]{
		\includegraphics[width=0.4\linewidth]{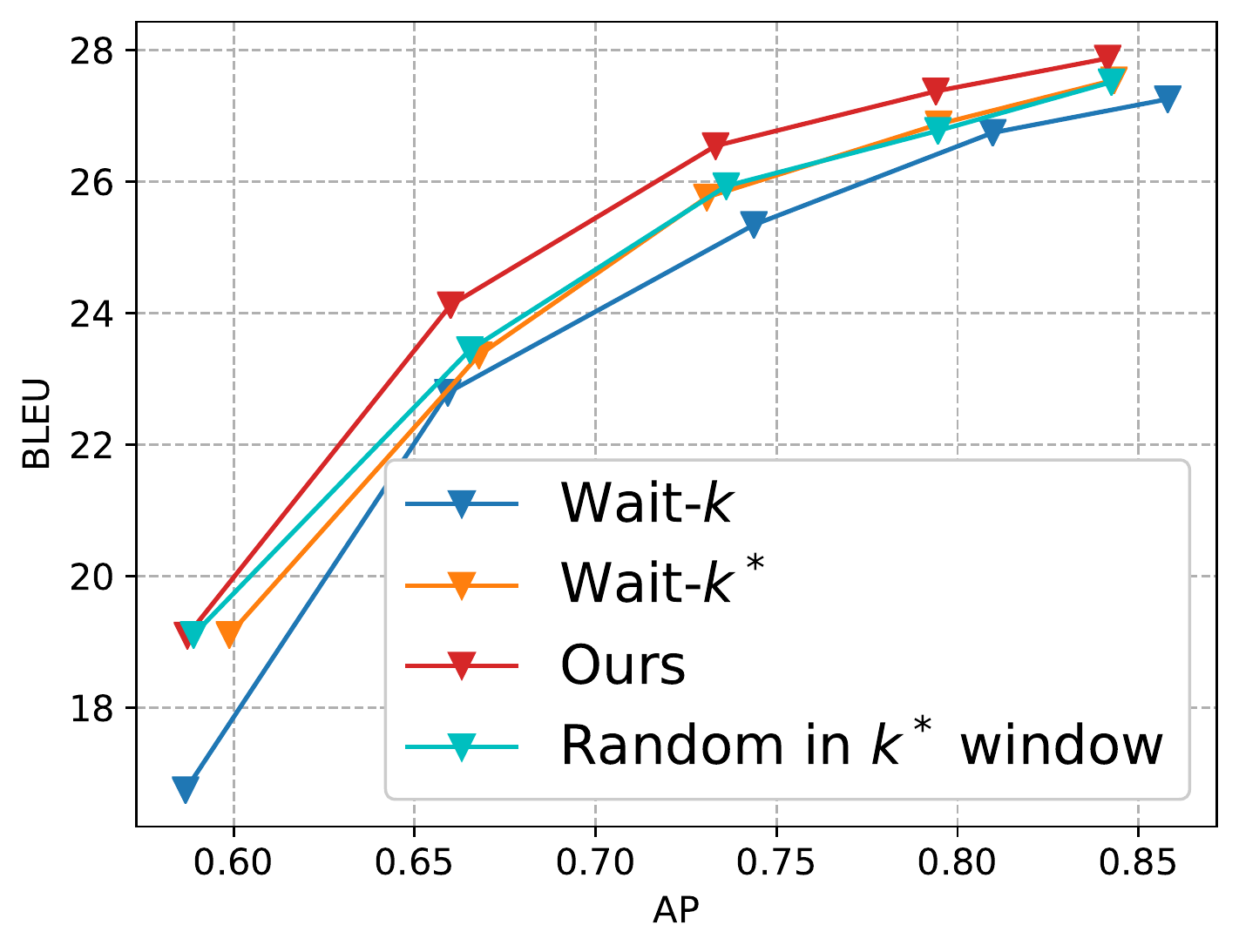}
	}
	\subfigure[BLEU-AL, IWSLT'14 En$\to$De]{
		\includegraphics[width=0.4\linewidth]{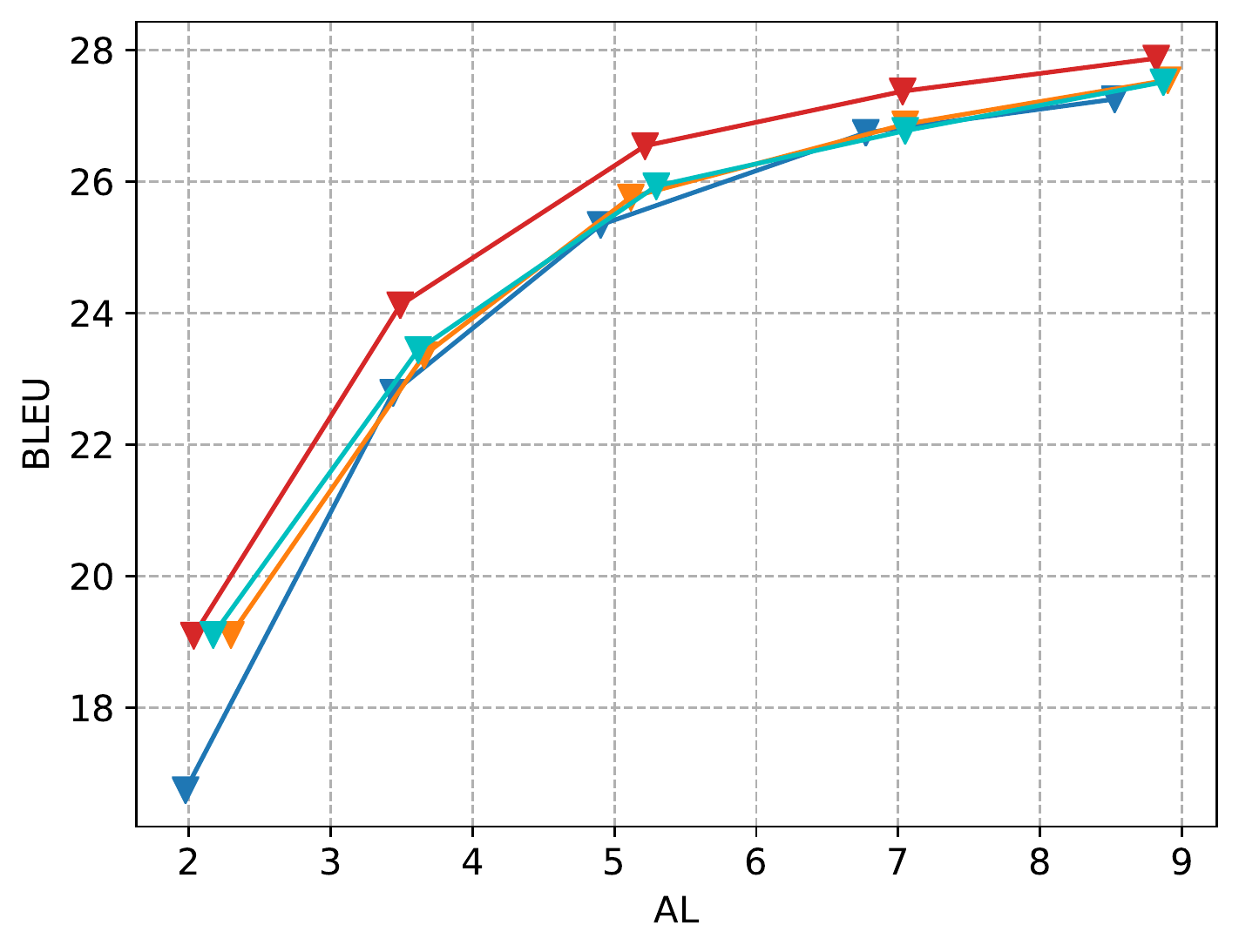}
	}
	\caption{Randomly selecting $k$ in a window around $k^*$.}
	\label{fig:random_around_kstar}
\end{figure}

\noindent(1) \textbf{Randomly selecting $k$ in a window around $k^*$}: We implement another baseline which is a combination of \waitkstar{} and MTL. After obtaining $k^*$, instead of sampling \waitm~on all possible $\{1,2,\cdots,M\}$, we sample on a smaller region around $k^*$. We conduct experiments on IWSLT En$\to$De and the results are in Figure~\ref{fig:random_around_kstar}. We can see that this variant achieves similar results with \waitkstar, but is not as good as our method. This shows the importance of using an adaptive controller to guide the training. 

\noindent(2) \textbf{A variant of CL}: Considering that the curriculum transfer learning is related to curriculum  learning, we implement self-paced learning (SPL) for \waitk{}, a CL method based on loss function: within each minibatch, we remove the $\tau\%$ sentences with the largest loss, where $\tau$ is gradually decreased from $40$ to $0$. On IWSLT En$\to$De, when $k=3$, the BLEU/AL/AP for SPL are $22.87$/$0.66$/$3.47$, which are worse than conventional CL ($23.41$/$0.66$/$3.48$). 

\begin{figure}[!htbp]
	\centering
	\subfigure[BLEU-AP, IWSLT'15 En$\to$Vi]{
		\includegraphics[width=0.4\linewidth]{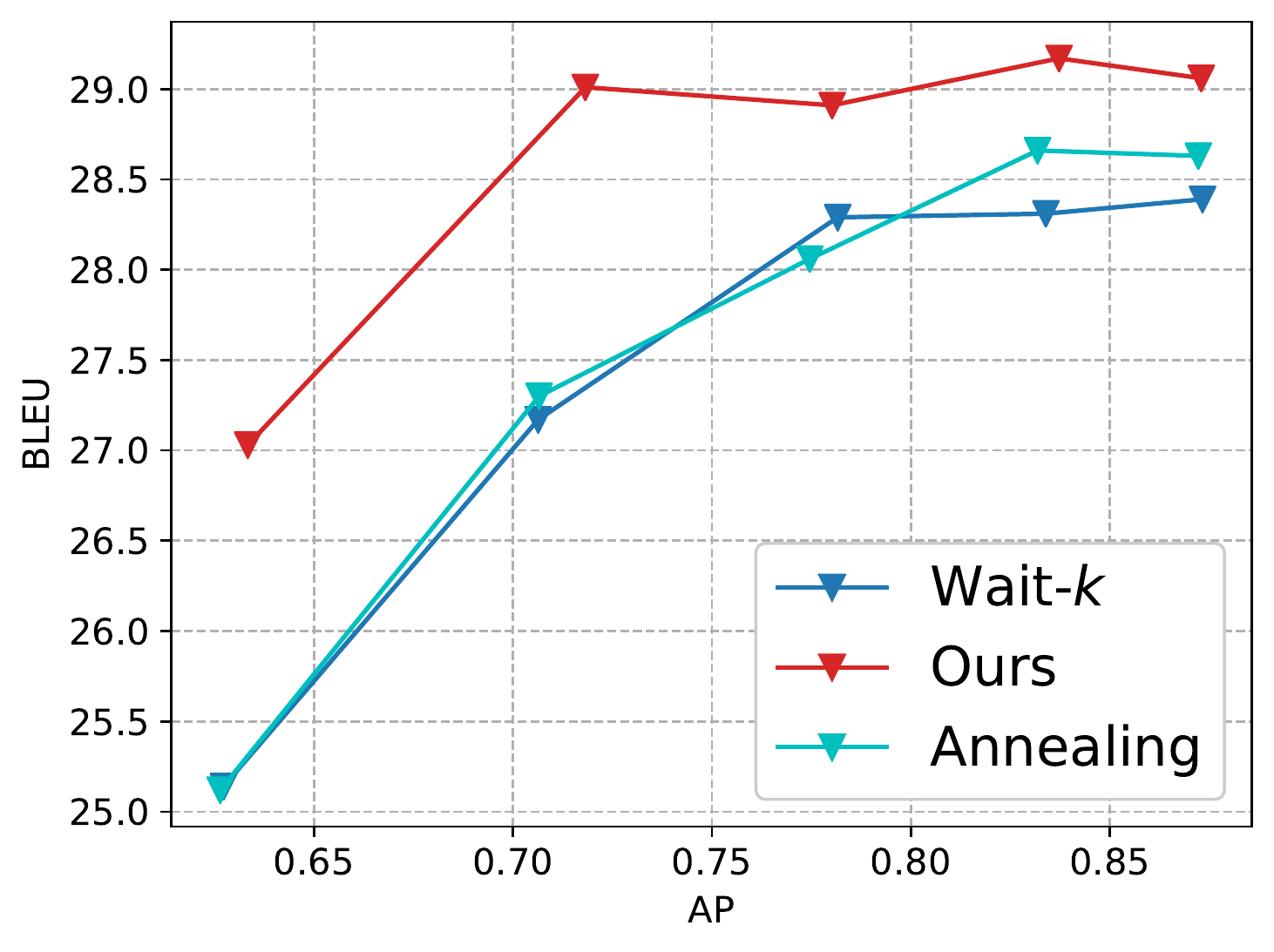}
	}
	\subfigure[BLEU-AL, IWSLT'15 En$\to$Vi]{
		\includegraphics[width=0.4\linewidth]{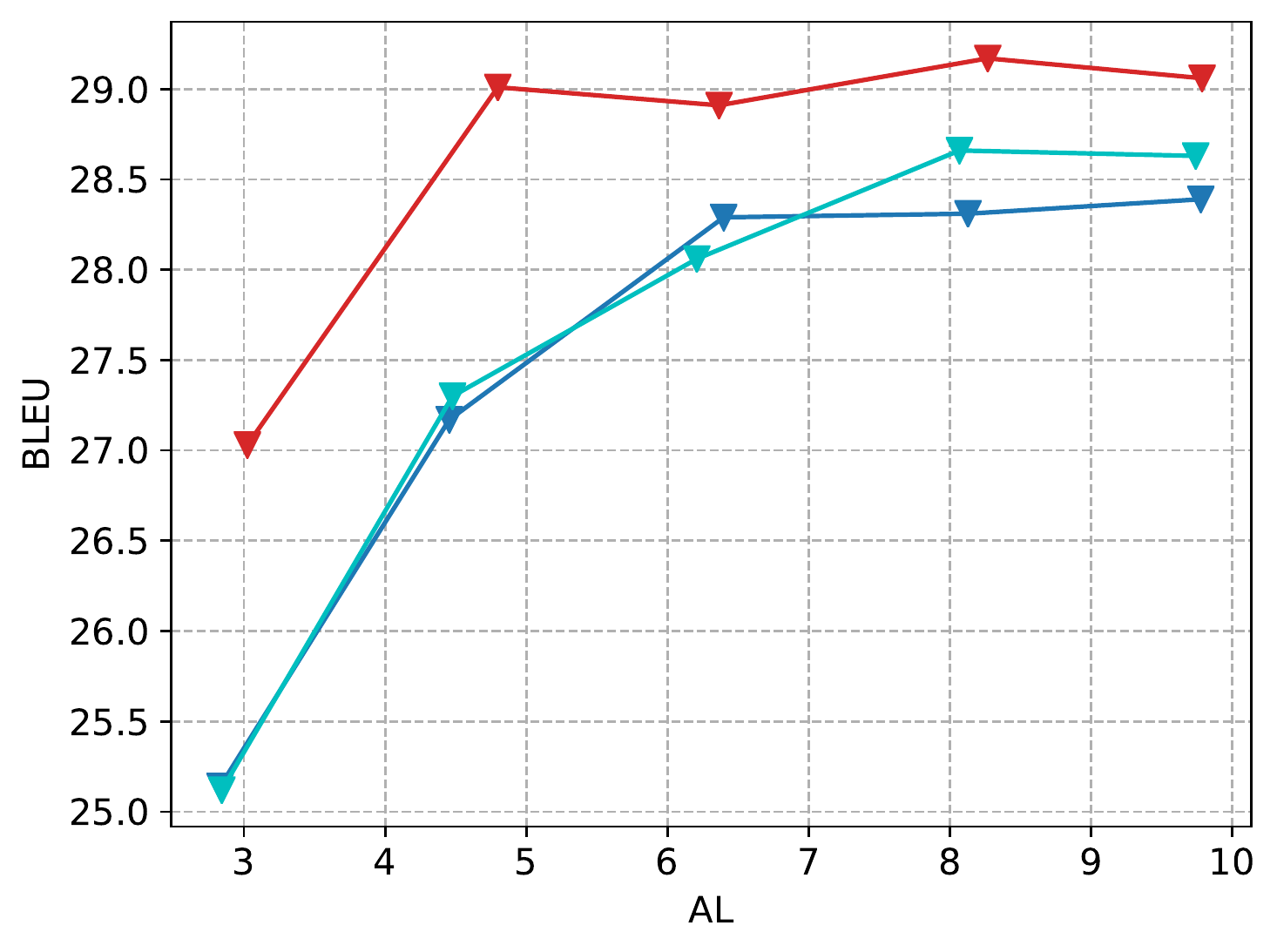}
	}
	\caption{The annealing strategy.}
	\label{fig:annealing}
\end{figure}

\noindent(3) \textbf{An annealing strategy}: Inspired by the fact that \waitkstar{} outperforms \waitk{}, we design a baseline where we randomly sample the waiting threshold $m$ from a distribution $p_t(m)$ at each training step $t$. The distribution $p_t(m)$ linearly anneals from a uniform distribution to a distribution which prefers larger $m$. We expect a single annealing strategy can train reasonably good models for different inference-time $k$ values.
Suppose the minimal $m$ value is $m_\textmd{min}$, the maximal $m$ value is $m_\textmd{max}$.  $m_\textmd{min}$ and $m_\textmd{max}$ are two integers and $m\in\{m_\textmd{min},m_\textmd{min}+1,\cdots,m_\textmd{max}\}$. The total training step is denote as $t_\text{max}$. $p_t(m)$ is mathematically defined as follows:
\begin{equation}
\begin{aligned}
p_t(m) &= (1 - \frac{t}{t_\text{max}}) \cdot p_\textmd{init}(m) + \frac{t}{t_\text{max}} \cdot p_\textmd{final}(m), \\
p_\textmd{init}(m) &= \frac{1}{m_\textmd{max} - m_\textmd{min} + 1},\quad p_\textmd{final}(m)=\frac{m}{\sum_{i=m_\textmd{min}}^{m_\textmd{max}}i}.
\end{aligned}
\end{equation}
The results on IWSLT'15 En$\to$Vi are shown in Figure~\ref{fig:annealing}, which shows this baseline brings limited improvement compared to \waitk{}. A possible reason is that this baseline cannot guarantee the best “annealing” strategy for each separate $k$, while our method can adaptively find the optimal strategy. Besides, as shown in Figure 5 in the main content, the learned strategies for different \waitk{} inference  are pretty different.

\section{More details about stock trend forecasting}
\subsection{Detailed introduction of the datasets}
For the Chinese A-share market dataset used in stock trend forecasting tasks, all the price and volume data are loaded from Qlib~\citep{yang2020qlib}. In this paper, only the stocks in CSI300 are included in the experiments. CSI300 consists of the 300 largest and most liquid A-share stocks, that can reflect the overall performance of China A-share market. 

\subsection{Training details}
For the stock trend forecasting models, we select the optimal hyper-parameters including learning rate, dropout rate, batch size and optimizer of scheduler and the sequence length $L$ on the validation set. We repeat each experiment by using $5$ random seed and report the average scores (RankIC, ICIR and MSE) to get the final result.

\subsection{Supplemental results}
We report the specific ICIR scores for stock trend forecasting task in Table~\ref{tab:icir}. Comparing with RankIC and MSE, similar phenomenon can be observed: 1) with temporally correlated tasks, our method significantly outperforms MTL and CL on most settings; 2) using a larger temporally correlated task set does not always bring improvements; 3) our method performs better than some strong models like GAT and lightGBM (briefly, LGB).

\begin{table}[htbp]
	\centering
	\begin{tabular}{cccc}
		\toprule
		\multicolumn{1}{c}{} & \multicolumn{1}{c}{\textbf{Task-1}} & \multicolumn{1}{c}{\textbf{Task-2}} & \multicolumn{1}{c}{\textbf{Task-3}} \\
		\midrule
		\textbf{GRU} & 0.373 & 0.127 & 0.111 \\
		\textbf{MLP} & 0.175 & 0.164 & 0.116 \\
		\textbf{LGB} & 0.297 & 0.180 & 0.070 \\
		\textbf{GAT} & 0.389 & 0.133 & 0.113 \\
		\midrule
		\multicolumn{4}{c}{\textbf{MTL}} \\
		\midrule
		\textbf{3tasks} & 0.467 & 0.179 & 0.083 \\
		\textbf{5tasks} & 0.444 & 0.146 & 0.115 \\
		\textbf{10tasks} & 0.403 & 0.134 & 0.135 \\
		\midrule
		\multicolumn{4}{c}{\textbf{CL}} \\
		\midrule
		\textbf{3tasks} & 0.397 & 0.198 & 0.112 \\
		\textbf{5tasks} & 0.419 & 0.200 & 0.110 \\
		\textbf{10tasks} & 0.393 & 0.200 & 0.117 \\
		\midrule
		\multicolumn{4}{c}{\textbf{Ours}} \\
		\midrule
		\textbf{3tasks} & 0.559 & 0.228 & 0.126 \\
		\textbf{5tasks} & 0.583 & 0.203 & 0.117 \\
		\textbf{10tasks} & 0.532 & 0.222 & 0.160 \\
		\bottomrule
	\end{tabular}%
	\caption{ICIR scores. The first four rows are the results of various model architectures without leveraging temporally correlated tasks. The following rows are the results with different algorithms and different task sets.}
	\label{tab:icir}%
\end{table}%

\end{document}